\documentclass[lettersize,journal]{IEEEtran}
\usepackage{amsmath,amsfonts}
\usepackage{algorithmic}
\usepackage{algorithm}
\usepackage{array}
\usepackage[caption=false,font=normalsize,labelfont=sf,textfont=sf]{subfig}
\usepackage{textcomp}
\usepackage{stfloats}
\usepackage{url}
\usepackage{verbatim}
\usepackage{graphicx}
\usepackage{cite}
\usepackage{xcolor}
\usepackage[table]{xcolor}
\usepackage{multirow}
\usepackage{makecell}
\usepackage{booktabs}
\usepackage{bm}
\usepackage{arydshln}
\usepackage{caption}
\begin{document}


\title{MonSter++: Unified Stereo Matching, Multi-view Stereo, and Real-time Stereo with Monodepth Priors}

\author{Junda Cheng$^*$, Wenjing Liao$^*$, Zhipeng Cai, Longliang Liu, Gangwei Xu, Xianqi Wang, Yuzhou Wang, Zikang Yuan, Yong Deng, Jinliang Zang, Yangyang Shi, Jinhui Tang,~\IEEEmembership{Member,~IEEE}, Xin Yang,~\IEEEmembership{Member,~IEEE}
\thanks{$^*$: Equal Contribution. Corresponding author: Xin Yang.}
\thanks{Junda Cheng, Wenjing Liao, Longliang Liu, Gangwei Xu, Xianqi Wang, Yuzhou Wang, and Xin Yang are with the School of Electronic Information and Communications,
Huazhong University of Science and Technology, Wuhan 430074, China (E-mail:\{jundacheng, wenjingliao, longliangl, gwxu, xianqiw, yuzhouwang, xinyang2014\}@hust.edu.cn).}
\thanks{Zhipeng Cai, Yangyang Shi are with Meta. (E-mail: czptc2h@gmail.com, shiyang1983@gmail.com)}
\thanks{Zikang Yuan is a postdoctoral fellow in AI Chip Center, Hong Kong University of Science and Technology (HKUST). (E-mail: zikangyuan@ust.hk).}
\thanks{Jinhui Tang is with the Nanjing Forestry University. (E-mail: tangjh@njfu.edu.cn).}
\thanks{Yong Deng, Jinliang Zang are with Autel Robotics. (E-mail: dengyong@autelrobotics.com, zangjinliang@autelrobotics.com)}
}



\maketitle

\begin{abstract}
We introduce MonSter++, a geometric foundation model for multi-view depth estimation, unifying rectified stereo matching and unrectified multi-view stereo. Both tasks fundamentally recover metric depth from correspondence search and consequently face the same dilemma: struggling to handle ill-posed regions with limited matching cues.
To address this, we propose MonSter++, a novel method that integrates monocular depth priors into multi-view depth estimation, effectively combining the complementary strengths of single-view and multi-view cues. MonSter++ fuses monocular depth and multi-view depth into a dual-branched architecture. Confidence-based guidance adaptively selects reliable multi-view cues to correct scale ambiguity in monocular depth. The refined monocular predictions, in turn, effectively guide multi-view estimation in ill-posed regions. This iterative mutual enhancement enables MonSter++ to evolve coarse object-level monocular priors into fine-grained, pixel-level geometry, fully unlocking the potential of multi-view depth estimation. MonSter++ achieves new state-of-the-art on both stereo matching and multi-view stereo. By effectively incorporating monocular priors through our cascaded search and multi-scale depth fusion strategy, our real-time variant RT-MonSter++ also outperforms previous real-time methods by a large margin. As shown in Fig.~\ref{fig:ranking}, MonSter++ achieves significant improvements over previous methods across eight benchmarks from three tasks—stereo matching, real-time stereo matching, and multi-view stereo, demonstrating the strong generality of our framework. Besides high accuracy, MonSter++ also demonstrates superior zero-shot generalization capability. We will release both the large and the real-time models to facilitate their use by the open-source community.
The code will be released at: \textcolor{magenta}{https://github.com/Junda24/MonSter-plusplus}.

\end{abstract}


\section{Introduction}

\IEEEPARstart{U}{nderstanding} 3D geometry from 2D images is a fundamental problem in computer vision and serves as the cornerstone of numerous real-world applications, such as autonomous driving, robotic navigation, and augmented reality. Among the many facets of 3D perception, absolute depth estimation plays a pivotal role, as it provides metric scene understanding that is crucial for downstream tasks involving interaction, planning, and decision-making.

\begin{figure}[t]
\centering
{\includegraphics[width=1.0\linewidth]{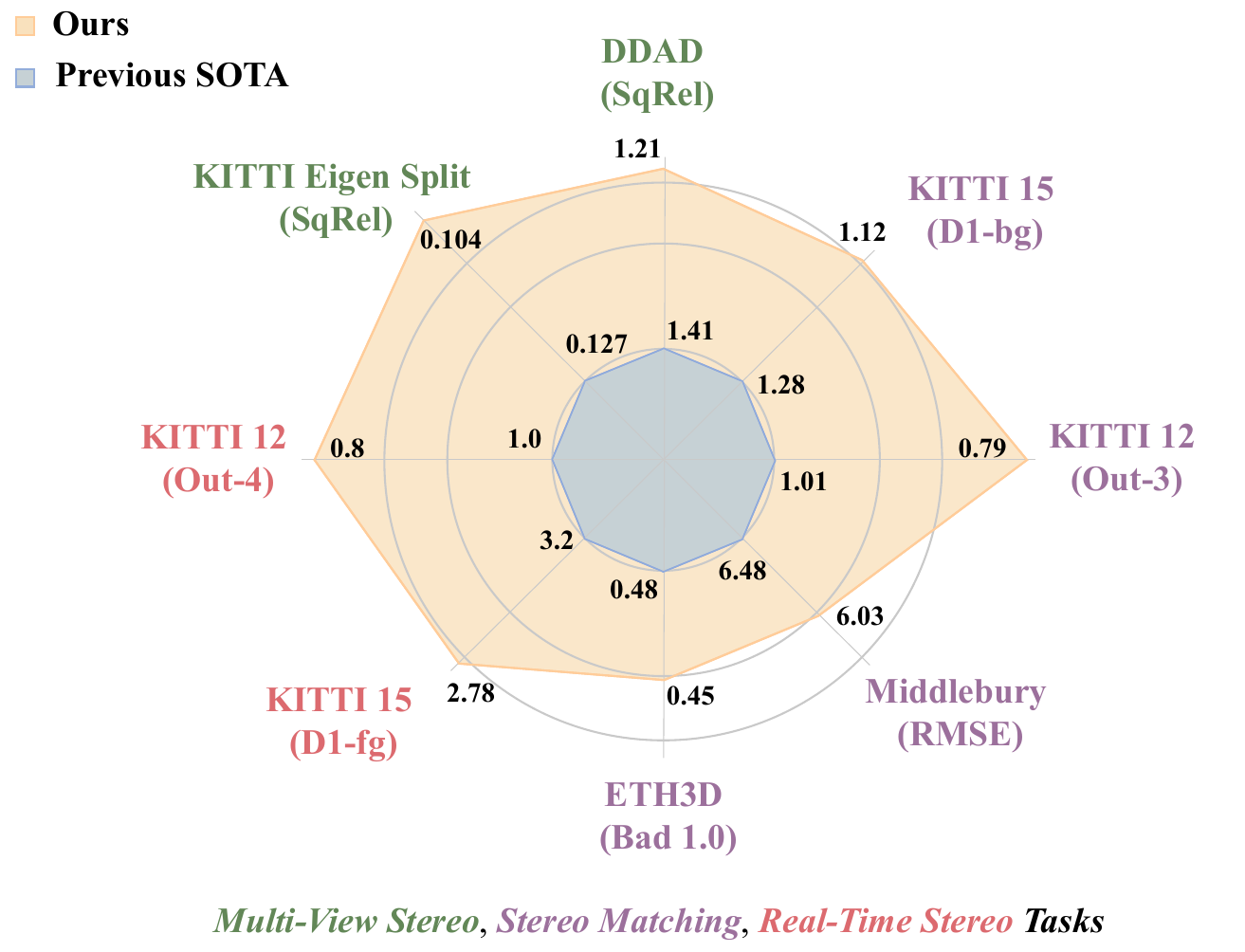}}
\caption{\textbf{The proposed MonSter++ vs previous SOTA on stereo matching, multi-view stereo (mvs), and real-time stereo.} MonSter++ demonstrates strong multi-task generalization, advanced previous SOTA by a large margin in all three tasks, and ranks $1^{st}$ across 8 leaderboards.}\label{fig:ranking}
\end{figure}

Obtaining accurate metric depth from multi-view data such as rectified stereo matching~\cite{psmnet, gwcnet, ganet, raft-stereo, selective-igev, igev, monster} and unrectified multi-view stereo~\cite{yao2018mvsnet, gu2020cascade, itermvs, mvs2d, mvsformer, afnet} has been widely adopted in practice. 
Both methods recover depth by establishing dense correspondences across multiple views along epipolar lines. In stereo matching, rectified image pairs constrain the epipolar lines to be horizontal, simplifying the correspondence search. In contrast, multi-view stereo handles images with arbitrary camera poses, where matching is performed along more general epipolar directions. In this paper, we adopt the term \textbf{multi-view depth estimation} as a unified formulation that encompasses both traditional stereo matching (two-view setting) and more general multi-view setups (three or more views).

With the rapid advancement of deep learning and large-scale datasets, learning-based methods~\cite{psmnet, gwcnet, raft-stereo, igev, selective-igev, monster, yao2018mvsnet, gu2020cascade, Mcstereo,itermvs, mvs2d, mvsformer, afnet, mvsformer++, zerostereo, xu2025banet, wang2025stereogen,wei2025wavelet} have dominated multi-view depth estimation and the corresponding leaderboards~\cite{eth3d, middlebury, dispNetC, kitti2015, ddad, kitti_eigen}. Mainstream multi-view depth estimation methods typically follow a four-stage pipeline: feature extraction, cost volume construction, cost aggregation, and depth regression. The most critical stages are cost volume construction and cost aggregation. While cost volume construction offers initial similarity estimates for potential correspondences across views, it is often prone to noise and ambiguity. Therefore, effective cost aggregation and regularization are crucial for refining the volume and achieving accurate, robust depth estimation. Based on different cost aggregation strategies, these methods can be roughly categorized into \emph{cost filtering-based} methods and \emph{iterative optimization-based} methods. Cost filtering-based methods~\cite{psmnet, gwcnet, acvnet, gcnet, aanet, dispNetC, unifying} construct 3D/4D cost volume using CNN features, followed by a series of 2D/3D convolutions for regularization and filtering to minimize mismatches. Iterative optimization-based methods~\cite{raft-stereo, crestereo, igev, dlnr} initially construct an all-pairs correlation volume, then index a local cost to extract motion features, which guide the recurrent units (ConvGRUs)~\cite{gru} to iteratively refine the depth map.



\begin{figure*}[t]
  \centering
      \includegraphics[width=0.99\linewidth,page=1]{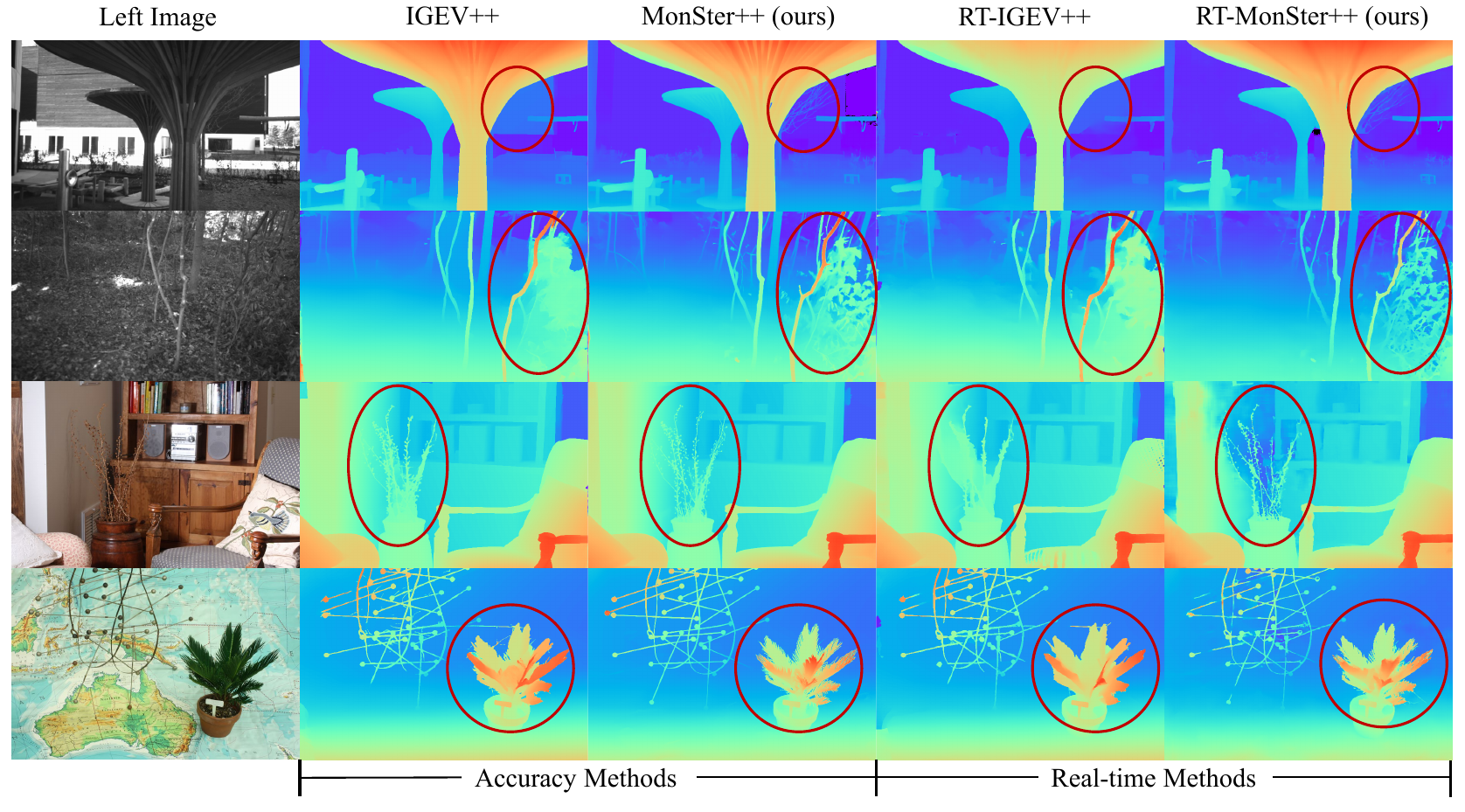}
    \caption{\textbf{Zero-shot generalization comparison on stereo matching benchmarks}. To validate the effectiveness of our method, we compare both our large model MonSter++ and the real-time variant RT-MonSter++ against state-of-the-art methods. Our method achieves particularly notable improvements on fine object boundaries in both accuracy models and real-time methods. Remarkably, RT-MonSter++ requires only 47 ms per inference yet even surpasses the accuracy-focused SOTA method IGEV++~\cite{igev++}, which takes 280 ms, particularly on edge regions. This highlights the superior efficiency of our method.}
    \label{fig:vis_rt_comp}
\end{figure*}

Both types of methods essentially derive depth through similarity matching, assuming the presence of visible correspondences across images. This poses challenges in \emph{ill-posed} regions with limited matching cues, e.g., occlusions, textureless areas, repetitive/thin structures, and distant objects with low pixel representation. Existing methods~\cite{psmnet, acvnet, segstereo, edgestereo, unifying, transformer, dlnr, transmvsnet, mvsformer, mvsformer++} address this issue by enhancing matching with stronger feature representations during feature extraction or cost aggregation. 
However, these methods still struggle to fundamentally resolve the issue of mismatching, limiting their practical performance.

\begin{figure}[ht]
\centering
{\includegraphics[width=0.96\linewidth]{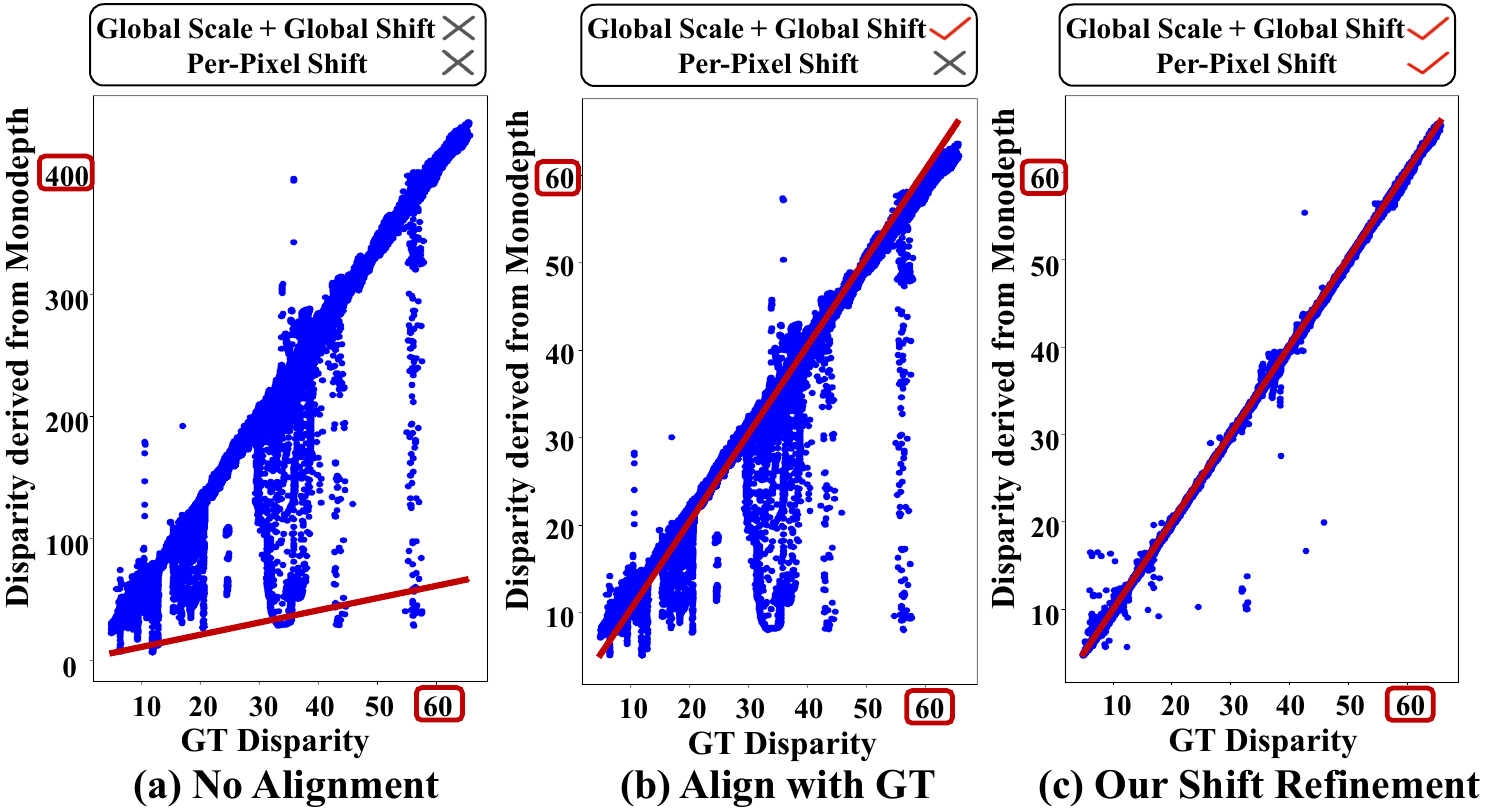}}
\caption{\textbf{Distance between GT disparity and the disparity derived from monocular depth~\cite{depthany_v2} on KITTI dataset.} The red line indicates identical disparity maps. (a): Without any alignment. (b) Align depth with GT using global scale and global shift values (same for all pixels). (c) The monocular disparity produced by MonSter++, with per-pixel shift refinement. Even globally aligned with GT, SOTA monocular depth models still exhibit severe noise. Our method MonSter++ effectively addresses this issue by refining monocular depth with per-pixel shift, which fully unlocks the power of monocular depth priors for stereo matching.}\label{fig:align_vis}
\end{figure}

Unlike multi-view depth estimation, monocular depth estimation directly recovers 3D from a single image, thereby avoiding the challenge of mismatching in ill-posed regions. While monocular depth provides complementary structural information for stereo, pre-trained models often yield relative depth with scale and shift ambiguities. As shown in Fig.\ref{fig:align_vis}, the prediction of monodepth models differs significantly from the ground truth. Even after global scale and shift alignment, substantial errors still persist, complicating pixel-wise fusion of monocular depth and multi-view depth. In this work, we formulate multi-view depth estimation as a problem of leveraging multi-view matching information to achieve per-pixel scale-shift recovery based on monocular depth estimation. We argue that this approach effectively combines the advantages of both single-view and multi-view algorithms while overcoming the limitations caused by the lack of matching cues.

To this end, we propose MonSter++, which constructs separate branches for single-view and multi-view depth estimation, and adaptively fuses them through \emph{Stereo Guided Alignment} (SGA) and \emph{Mono Guided Refinement} (MGR) modules. SGA initially rescales the monocular depth to a coarse metric scale via global alignment with multi-view depth. Then it uses condition-guided GRUs to adaptively select reliable stereo cues for updating the per-pixel monocular depth shift. Symmetric to SGA, MGR uses the optimized monocular depth as the condition to adaptively refine the multi-view depth in regions where matching fails. Through multiple iterations, the two branches effectively complement each other: 1) Though beneficial at coarse object-level, directly and unidirectionally fusing monodepth into multi-view stereo suffers from scale-shift ambiguities, which often introduces noise in complex regions such as slanted or curved surfaces. Refining monodepth with stereo resolves this issue effectively, ensuring the robustness of MonSter++. 2) The refined monodepth in turn provides strong guidance to multi-view depth in challenging regions. E.g., the perception ability of multi-view depth degrades with distance due to smaller pixel proportions and the increased matching difficulty. Monodepth models pre-trained on large-scale datasets are less affected by such issues, which can effectively improve multi-view depth in the corresponding region.

By fully leveraging the strengths of monocular depth priors, MonSter++ achieves state-of-the-art performance in both stereo matching and multi-view stereo, ranking first on four most commonly used stereo matching benchmarks ~\cite{eth3d, middlebury, kitti2012, kitti2015} and leading two key multi-view depth benchmarks~\cite{ddad, kitti_eigen}. In addition to improving accuracy, MonSter++ substantially enhances the generalization ability of stereo matching, achieving competitive results on real-world scenarios with minimal synthetic training data and surpassing existing state-of-the-art methods. To fully leverage the advantages of our depth fusion architecture, we also introduce a real-time version of MonSter++, named RT-MonSter++, offering an attractive solution for time-constrained applications. By leveraging monocular priors as in MonSter++ and adopting an efficient cascaded search and multi-scale depth fusion strategy, RT-MonSter++ sets a new state of the art among real-time methods on the KITTI benchmarks, while also delivering strong generalization. Notably, RT-MonSter++ even surpasses the accuracy-oriented SOTA method IGEV++, particularly on thin structures, as shown in Fig.~\ref{fig:vis_rt_comp}. Its balance of accuracy and speed makes it highly practical for deployment on mobile platforms like drones, autonomous vehicles, and robotic systems.

\begin{figure*}[t]
  \centering
    \includegraphics[width=0.95\linewidth,page=1]{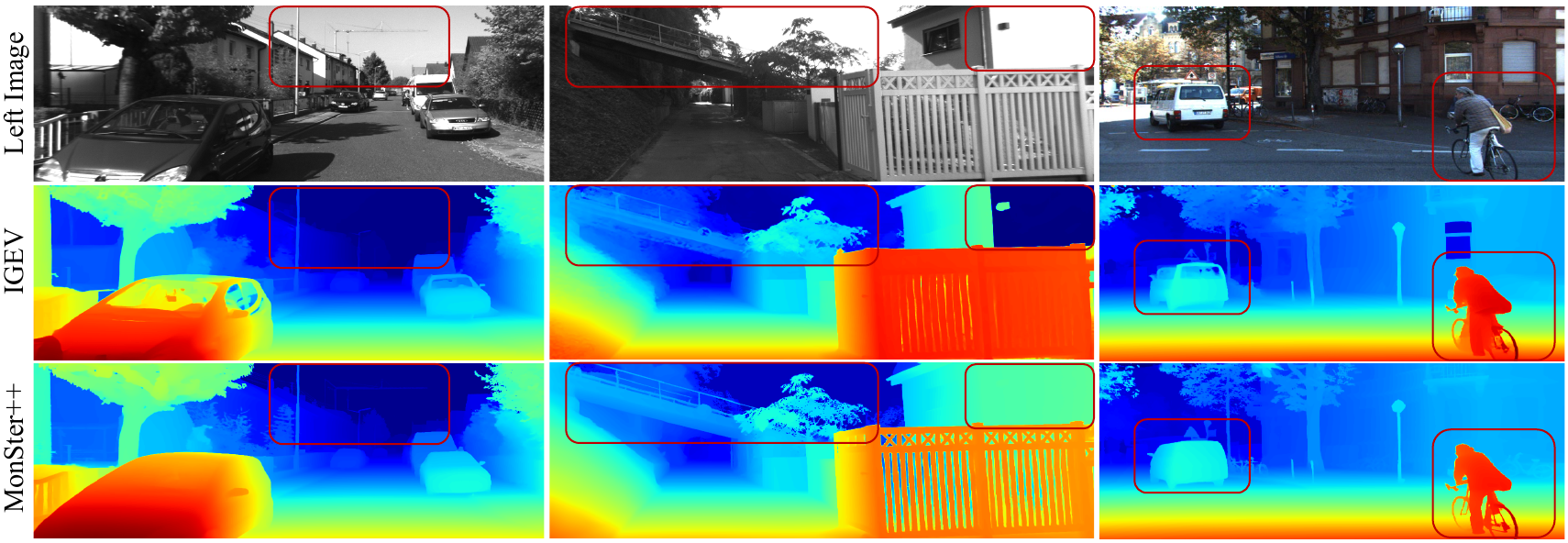}
    \caption{\textbf{Zero-shot generalization comparison}: all models are trained on Scene Flow and tested directly on KITTI. Compared to the baseline IGEV~\cite{igev}, our method MonSter++ shows significant improvement in challenging regions such as reflective surfaces, textureless areas, fine structures, and distant objects.}
    \label{fig:vis_kitti}
\end{figure*}

Experimental results show that our method significantly outperforms previous SOTA methods across all three tasks—stereo matching, multi-view stereo, and real-time stereo matching—highlighting the efficiency and versatility of our framework. As shown in Fig.~\ref{fig:ranking}, MonSter++ ranks $1^{st}$ across 4 widely-used stereo matching leaderboards: KITTI 2012 \cite{kitti2012}, KITTI 2015 \cite{kitti2015}, Middlebury \cite{middlebury} and ETH3D \cite{eth3d}. It advances SOTA by up to $21.8\%$ compared to the previous best method on KITTI 2012. As for multi-view stereo task, MonSter++ also ranks $1^{st}$ across 2 widely-used multi-view depth benchmarks: DDAD~\cite{ddad} and KITTI Eigen split~\cite{kitti_eigen}. By leveraging a coarse-to-fine framework, we extend MonSter++ into a more efficient and lightweight variant, RT-MonSter++, which achieves real-time inference while surpassing all existing real-time methods on the KITTI benchmark in accuracy. RT-MonSter++ outperforms the previous best real-time method RT-IGEV by 20.0\% on the KITTI 2012 benchmark. 
In addition, by effectively leveraging monocular priors, both MonSter++ and RT-MonSter++ exhibit superior zero-shot generalization. Under the same training setup, our methods achieve the highest zero-shot performance on KITTI 2012, KITTI 2015, ETH3D, and Middlebury datasets. Notably, RT-MonSter++ requires only 47 ms for inference to reach performance comparable to the accuracy-oriented model IGEV (180 ms), and even surpasses it on three out of the four datasets. Our \textbf{contributions} can be summarized below:
\begin{itemize}
    \setlength{\parskip}{0pt}
    \setlength{\parsep}{0pt}
    \item We propose a unified framework MonSter++ for multi-view depth estimation that handles both rectified and unrectified input images. MonSter++ fully leverages the pixel-level monocular depth priors to significantly improve the accuracy in ill-posed regions and fine structures. As shown in Fig.~\ref{fig:ranking}, MonSter++ ranks $1^{st}$ across 4 widely-used stereo matching leaderboards and 2 mainstream multi-view stereo benchmarks.

    \item We propose RT-MonSter++, which achieves real-time inference while delivers the best accuracy among all published real-time methods on the KITTI 2012~\cite{kitti2012} and 2015~\cite{kitti2015} benchmarks. 
    
    \item We perform large scale training on 2M image pairs collected from public datasets, enabling superior
    zero-shot generalization over previous methods consistently across diverse datasets for both MonSter++ and RT-MonSter++. 
    All models and training code will be released to the community.   
\end{itemize}

\section{Related Work}
\label{sec:related_work}
Depth estimation is a key technology in 3D perception, crucial for downstream tasks such as 3D reconstruction~\cite{mvs2d, afnet, mvsformer, L-magic, mvsformer++}, localization~\cite{D2VO, CR-LDSO, stvo, romeo, PriOr-Flow, LiDAR-Inertial, Voxel-svio, Sr-lio, SR-LIVO, SDV-LOAM}, and decision-making. Existing visual depth estimation methods can be broadly categorized into monocular depth estimation, stereo matching, and multi-view stereo. We first review related work of each task independently, and then discuss how recent studies unify them by incorporating monocular depth priors and/or semantic cues into multi-view depth estimation.

\subsection{Monocular depth estimation} 
Monocular depth estimation is a fundamental problem in computer vision. In recent years, CNN-based methods~\cite{yin2021virtual, bhat2021adabins, yuan2022new, Yin2019enforcing, li2023, li2023learning} have gained considerable attention, typically formulating the task as either a per-pixel classification~\cite{bhat2021adabins, fu2018deep, Yin2019enforcing} or regression problem~\cite{lee2019big, hao2018detail, laina2016deeper, xu2018structured}. To enhance performance, some approaches focus on improving feature representations~\cite{yuan2022new, Qi_2018_CVPR}, while others propose more effective loss functions~\cite{bhat2021adabins, yin2021virtual, Yin2019enforcing} or utilize mixed-data training strategies~\cite{yin2020diversedepth, yin2022towards, eftekhar2021omnidata}. Recent monocular depth estimation methods~\cite{depthany_v1, depthany_v2, wang2025moge2accuratemonoculargeometry, prompting}, built upon powerful image backbones~\cite{dinov2}, scale up both model and dataset sizes by leveraging a mix of real and synthetic data, achieving strong generalization across diverse scenes.
Although these methods have achieved significant progress on benchmark datasets, their accuracy still lags behind that of multi-view geometry-based approaches. In our system, we incorporate a monocular depth estimation module due to its robustness in low-texture regions and dynamic scenes to help improve overall performance and adaptability.

\subsection{Stereo matching} 
Mainstream stereo matching methods recover disparity from matching costs. These methods can generally be divided into two categories: cost filtering-based methods and iterative optimization-based methods. Cost filtering-based methods~\cite{gwcnet, dispNetC, unifying,coatrsnet, RSSM, gu2020cascade, cfnet} typically construct a 3D/4D cost volume using feature maps and subsequently employ 2D/3D CNN to filter the volume and derive the final disparity map. Constructing a cost volume with strong representational capacity and accurately regressing disparity from a noisy cost volume are key challenges. \cite{psmnet} proposes a stacked hourglass 3D CNN for better cost regularization. \cite{gwcnet, acvnet, pcwnet} propose the group-wise correlation volume, the attention concatenation volume, and the pyramid warping volume respectively to improve the expressiveness of the matching cost. Recently, a novel class of methods based on iterative optimization~\cite{raft-stereo, crestereo, igev, dlnr, bartolomei2024stereo, selective-igev} has achieved SOTA performance in both accuracy and efficiency. These methods use ConvGRUs to iteratively update the disparity by leveraging local cost values retrieved from the all-pairs correlation volume. Similarly, iterative optimization-based methods also primarily focus on improving the matching cost construction and the iterative optimization stages. CREStereo \cite{crestereo} proposes a cascaded recurrent network to update the disparity field in a coarse-to-fine manner. IGEV \cite{igev} proposes a Geometry Encoding Volume that encodes geometry and context information for a more robust matching cost. Both types of methods essentially derive depth from matching costs, which are inherently limited by ill-posed regions. 


\subsection{Multi-View stereo}
Numerous methods have been developed to estimate depth from multi-view images under the assumption of known camera intrinsics and extrinsics. Early work such as \cite{zbontar2015computing} pioneers the integration of deep feature learning into the multi-view stereo pipeline, although it still relies on traditional cost aggregation techniques. Later, \cite{yao2018mvsnet} introduces a breakthrough approach by constructing a differentiable cost volume, followed by 3D convolutional neural networks (CNNs) for regularization, significantly improving depth estimation performance. Building upon this framework, \cite{long2021multi} proposes a hybrid cost regularization network that decouples 3D local matching and 2D global context modeling to reduce computational overhead. Subsequent methods \cite{yao2019recurrent, yan2020dense, cai2023riavmvs} further enhance performance and robustness through iterative residual refinement, while others \cite{chen2019point, yu2020fast, gu2020cascade} adopt coarse-to-fine strategies to balance accuracy and efficiency. More recently, transformer-based architectures \cite{transmvsnet, liao2022wtmvsnetwindowbasedtransformersmultiview, mvsformer++} have been explored to enhance global context modeling for more robust feature representations. For example, MVSFormer \cite{mvsformer} with hierarchical ViTs delivers substantial improvements over conventional pyramid convolution-based approaches. MVS2d\cite{mvs2d} proposes an attention mechanism that aggregates features along the epipolar line for each query pixel in the reference image, significantly improving computational efficiency. Furthermore, MVSFormer++\cite{mvsformer++} introduces customized attention mechanisms tailored to different MVS modules, enabling better module adaptation. Despite these advances, the effectiveness of current MVS methods often hinges on scenes with large parallax and rich textures, and they struggle with dynamic content. Moreover, these approaches typically require precise relative pose estimation between views and are sensitive to errors in camera pose, limiting their robustness in practical applications.

\subsection{Multi-view depth estimation with Structural Priors} 
Multi-view depth estimation in ill-posed regions remains a significant challenge. To address this, prior work—both in stereo matching and multi-view stereo—has explored the integration of structural cues. For instance, EdgeStereo \cite{edgestereo} incorporates edge detection features into the disparity estimation pipeline to enhance performance around object boundaries, while SegStereo \cite{segstereo} employs semantic segmentation cues to improve matching in textureless areas. However, such cues primarily offer object-level priors and lack the granularity required for accurate pixel-level depth estimation. As a result, these methods often struggle in scenes containing large curved or slanted surfaces. To provide finer structural guidance, several approaches have introduced monocular depth estimation as a complementary prior. CLStereo \cite{monocular} incorporates a monocular branch to transfer geometric context into the stereo network, acting as a regularizer during matching. \cite{facil2017single} significantly improves performance by fusing fine-grained local structures extracted from single-view methods with robust multi-view cues, particularly in regions with large parallax and strong gradients. \cite{bae2022multi} first predicts an initial depth map using a monocular depth estimation network, and then constructs a lightweight cost volume centered around the initial depth, thereby achieving a balance between accuracy and efficiency. Similarly, Los \cite{los} utilizes monocular depth to generate local slanted planes that guide disparity refinement. FoundationStereo~\cite{foundationstereo} employs a side-tuning feature backbone to adapt monocular priors from DepthAnythingV2 \cite{depthany_v2} to stereo inputs, thereby enhancing its generalization capability. AFNet~\cite{afnet} further enhances depth estimation performance and robustness through confidence-based adaptive fusion of predictions from the multi-view and single-view branches. In summary, all the aforementioned methods use monocular depth merely as prior. However, monocular depth is inherently affected by scale and shift ambiguities, as illustrated in Fig.~\ref{fig:align_vis}, which limits its effective utilization and can introduce substantial noise when employed directly as a prior. To address this, our MonSter++ framework leverages monocular depth in a more robust manner by adaptively selecting reliable stereo disparities to calibrate the scale and shift of the monocular predictions. This design effectively integrates dense geometric priors while suppressing erroneous signals, leading to significantly improved performance in ill-posed regions.

\begin{figure*}[t]
    \centering
    \includegraphics[width=0.90\linewidth]{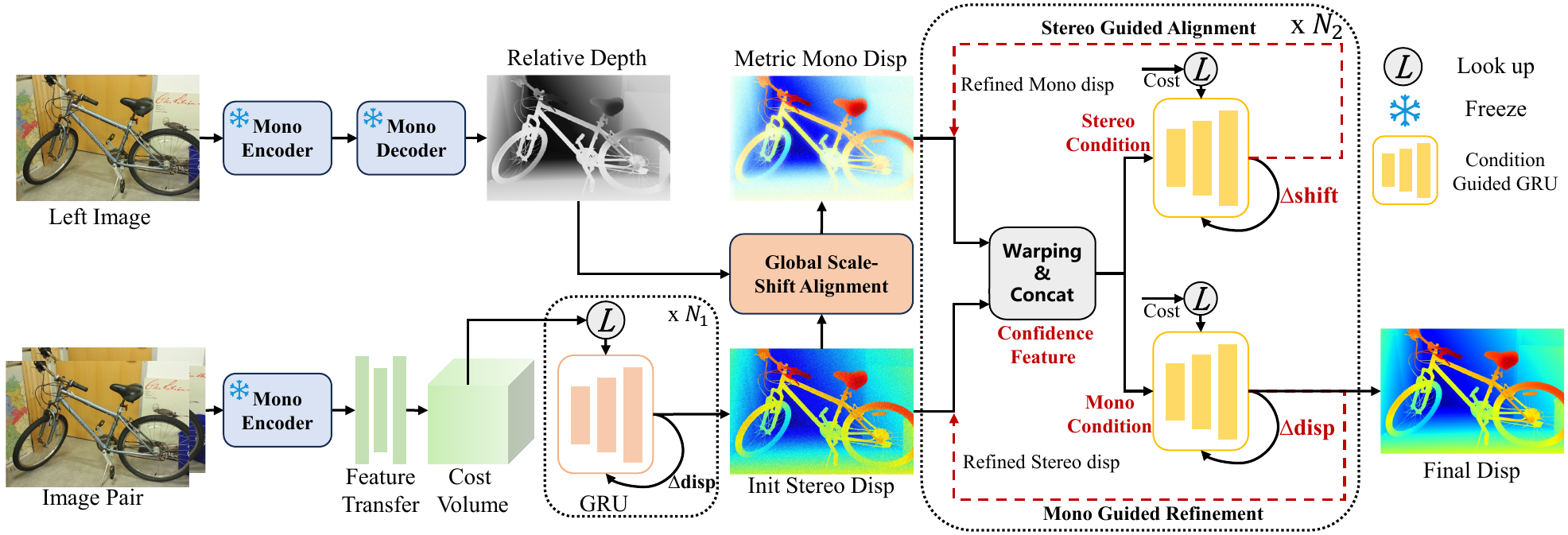}
    \caption{\textbf{Overview of MonSter++.} MonSter++ consists of a monocular depth estimation branch, a stereo matching branch, and a mutual refinement module. It iteratively improves one branch with priors from the other, effectively resolving the ill-posedness in stereo matching.}
    \label{fig:network}
\end{figure*}

\section{Method}
\label{sec:method}

We propose MonSter++, a unified framework that can be applied to both stereo matching and multi-view stereo for accurate metric depth estimation. For both tasks, the network architecture remains identical, with the only difference in the cost volume construction—a parameter-free process that operates by directly measuring feature similarity. Therefore, we first detail the MonSter++ framework using stereo matching as an example (Sec.~\ref{sec:stereo_matching}), then explicitly contrast it with the multi-view stereo model (Sec.~\ref{sec:mvs}). In addition, we describe details of our real-time version of MonSter++, named RT-MonSter++ (Sec.~\ref{sec:RT_monster}).

\subsection{Stereo matching}\label{sec:stereo_matching}

As shown in Fig.~\ref{fig:network}, MonSter++ consists of 1) a \emph{monocular depth branch}, 2) a \emph{stereo matching branch}, and 3) a \emph{mutual refinement module}. The two branches estimate initial monocular depth and the stereo disparity, which are fed into mutual refinement to iteratively improve each other.

\subsubsection{Monocular and Stereo Branches}
\label{sec:2branch}

The monocular depth branch can leverage most monocular depth models to achieve non-trivial performance improvements (see Sec.\ref{sec:ablation} for analysis). 
Our best empirical configuration uses pretrained DepthAnythingV2~\cite{depthany_v2} as the monocular depth branch, which uses DINOv2~\cite{dinov2} as the encoder and DPT~\cite{DPT} as the decoder. The stereo matching branch follows IGEV~\cite{igev} to obtain the initial stereo disparity, with modifications only to the feature extraction component, as shown in Fig.~\ref{fig:network}. To efficiently and fully leverage the pretrained monocular model, the stereo branch shares the ViT encoder in DINOv2 with the monocular branch, with parameters frozen to prevent the stereo matching training from affecting its generalization ability. Moreover, the ViT architecture extracts feature at a single resolution, while recent stereo matching methods commonly utilize multi-scale features at four scales (1/32, 1/16, 1/8, and 1/4 of the original image resolution). To fully align with IGEV, we employ a stack of 2D convolutional layers, denoted as the feature transfer network, to downsample and transform the ViT feature into a collection of pyramid features $\mathcal{F}=\{\boldsymbol{F}_{0},\boldsymbol{F}_{1},\boldsymbol{F}_{2}, \boldsymbol{F}_{3}\}$, where $\boldsymbol{F}_{k}\in\mathbb{R}^{\frac{H}{2^{5-k}} \times \frac{W}{2^{5-k}} \times {c}_k}$. 
We follow IGEV to construct the Geometry Encoding Volume and use the same ConvGRUs for iterative optimization. To balance accuracy and efficiency, we perform only $N_{1}$ iterations to obtain an initial stereo disparity with reasonable quality.

\subsubsection{Mutual Refinement}
Once the initial (relative) monocular depth and stereo disparity are obtained, they are fed into the mutual refinement module to iteratively refine each other. We first perform \emph{global scale-shift alignment}, which converts monocular depth into a disparity map and aligns it coarsely with stereo outputs. Then we iteratively perform a dual-branched refinement: \emph{Stereo guided alignment (SGA)} leverages stereo cues to update the per-pixel shift of the monocular disparity; \emph{Mono guided refinement (MGR)} leverages the aligned monocular prior to further refine stereo disparity.

\textbf{Global Scale-Shift Alignment.} Global Scale-Shift Alignment performs least squares optimization over a global scale $s_G$ and a global shift $t_{G}$ to coarsely align the inverse monocular depth with the stereo disparity: 
\begin{equation}
\begin{aligned}
s_G, t_G = & \; \arg \min _{s_G, t_G} \sum\limits_{i \in \Omega } \left(s_G \bm{D}_{M}(i)+t_G-\bm{D}_{S}^0(i)\right)^{2} \\
&  \;\bm{D}_{M}^0 = s_G \bm{D}_{M}+t_G,
\end{aligned}
\label{equ:align1}
\end{equation}
where $\bm{D}_{M}(i)$ and $\bm{D}^0_{S}(i)$ are the inverse monocular depth and stereo disparity at the i-th pixel. $\Omega$ represents the region where stereo disparity values fall between 20\% to 90\% sorting in ascending order, which helps to filter unreliable regions such as the sky, extremely distant areas, and close-range outliers. Intuitively, this step converts the inverse monocular depth into a disparity map coarsely aligned with the stereo disparity, enabling effective mutual refinement in the same space. We call this aligned disparity map $\bm{D}_M^0$ the \emph{monocular disparity} in the remainder of the paper.

\textbf{Stereo Guided Alignment (SGA).} Though coarsely aligned, a unified shift $t_G$ is not sufficient to recover accurate monocular disparity at different pixels. To fully release the potential of monocular depth prior, SGA leverages intermediate stereo cues to further estimate a per-pixel residual shift. To avoid introducing noisy stereo cues, SGA uses \emph{confidence based guidance}. In each update step $j$, 
we compute the confidence using the flow residual map $\bm{F}_{S}^j$, which is obtained by warping and subtracting features based on the stereo disparity $\bm{D}_{S}^j$ as:
\begin{equation}
\begin{aligned}
\bm{F}_{S}^j(x,y) = \left\|  \bm{F}_{3}^L(x,y) - \bm{F}_{3}^R(x-\bm{D}_S^j,y) \right\| _{1}
\end{aligned}
\label{equ:warp}
\end{equation}
where $\boldsymbol{F}_{3}^L$,$\boldsymbol{F}_{3}^R$ represent the quarter-resolution features of left and right images in $\mathcal{F}$ respectively. For each iteration, we also use the current stereo disparity $\bm{D}_S^j$ to index from the Geometry Encoding Volume to obtain geometry features of stereo branch $\bm{G}_{S}^j$ follow IGEV. Concatenated with $\bm{F}_{S}^j$ and $\bm{D}_S^j$, we obtain the stereo condition as:
\begin{equation}
\begin{aligned}
x_{S}^j = & \; [\text{En}_{g}([\bm{G}_{S}^j, \bm{F}_{S}^j, \bm{D}_S^j]),\text{En}_{d}(\bm{D}_{M}^j), \bm{D}_{M}^j],\\
\end{aligned}
\end{equation}
where $\text{En}_{g}$ and $\text{En}_{d}$ are two convolutional layers for feature encoding. We feed $x_{S}^j$ into condition-guided ConvGRUs to update the hidden state $h_{m}^{i-1}$ of monocular branch as:
\begin{equation}
\begin{aligned}
z^j = & \;\sigma(\text{Conv}([h_{M}^{j-1}, x_{S}^j], W_z) + c_k), \\
r^j = & \;\sigma(\text{Conv}([h_{M}^{j-1}, x_{S}^j], W_r) + c_r), \\
\Tilde{h}_{M}^j = & \,\tanh(\text{Conv}([r^j \odot h_{M}^{j-1}, x_{S}^j], W_h) + c_h), \\
h_{M}^j = & \;(1-z^j) \odot h_{M}^{j-1} + z^j \odot \Tilde{h}_{M}^j,
\end{aligned}
\label{equ:gru}
\end{equation}
where $c_k$, $c_r$, $c_h$ are context features. Based on the hidden state $h_{M}^{j}$, we decode a \emph{residual shift} $\Delta \bm{t}$ through two convolutional layers to update the monocular disparity:
\begin{equation}
\begin{aligned}
\bm{D}_{M}^{j+1} = & \; \bm{D}_{M}^j + \Delta \bm{t}.
\end{aligned}
\label{equ:update}
\end{equation}

\textbf{Mono Guided Refinement (MGR).} Symmetric to SGA, MGR leverages the aligned monocular disparity to address stereo deficiencies in ill-posed regions, thin structures, and distant objects. Specifically, we employ the same condition-guided GRU architecture with independent parameters to refine stereo disparity. We simultaneously calculate the flow residual maps $\bm{F}_{M}^j$, $\bm{F}_{S}^j$ and geometric features $\bm{G}_{M}^j$, $\bm{G}_{S}^j$ for both the monocular and stereo branches, providing a comprehensive stereo refinement guidance:
\begin{equation}
\begin{aligned}
& \;\bm{F}_{M}^j(x,y) =  \left\|  \boldsymbol{F}_{3}^L(x,y) - \boldsymbol{F}_{3}^R(x-\bm{D}_{M}^j,y) \right\| _{1}, \\
&\; x_M^j = [\text{En}_{g}([\bm{G}_{M}^j, \bm{F}_{M}^j, \bm{D}_{M}^j]),\text{En}_{d}(\bm{D}_{M}^j), \bm{D}_{M}^j, \\
& \; \text{En}_{g}([\bm{G}_{S}^j, \bm{F}_{S}^j, \bm{D}_S^j]),\text{En}_{d}(\bm{D}_{S}^j), \bm{D}_{S}^j],\\
\end{aligned}
\end{equation}
\noindent where $\bm{G}_M^j$ is the geometry features of the monocular branch obtained by indexing from the Geometry Encoding Volume using the monocular disparity $\bm{D}_{M}^j$. $x_M^j$ represents the monocular condition for ConvGRUs, and we use Eq.~\eqref{equ:gru} to update the hidden state $h_{S}^{i}$ similarly, with only the condition input changed from $x_S^i$ to $x_M^i$. We use the same two convolutional layers to decode the residual disparity $\triangle \bm{d}$ and update the current stereo disparity following Eq.~\eqref{equ:update}.
After $N_{2}$ rounds of dual-branched refinement, the disparity of the stereo branch is the final output of MonSter++.

\subsubsection{Loss Function}
We use the L1 loss to supervise the output from two branches. We denote the set of disparities from the first $N_{1}$ iterations of the stereo branch as $\{\bm{d}_{i}\}_{i=0}^{{N}_{1} -1}$ and follow \cite{raft-stereo} to exponentially increase the weights as the number of iterations increases. The total loss is defined as the sum of the monocular branch loss $\mathcal{L}_{Mono}$ and the stereo branch loss $\mathcal{L}_{Stereo}$ as follows:
\begin{equation}
\begin{aligned}
     \mathcal{L}_{Stereo} = & \; \sum_{i=0}^{{N}_{1}-1} \gamma^{{N}_{1}+{N}_{2}-i} ||\bm{d}_i-\bm{d}_{gt}||_1 + \\
    & \;\sum_{i={N}_{1}}^{{N}_{1}+{N}_{2}-1} \gamma^{{N}_{1}+{N}_{2}-i} ||\bm{D}_{S}^{i-{N}_{1}}-\bm{d}_{gt}||_1,\\
    \mathcal{L}_{Mono} = & \; \sum_{i={N}_{1}}^{{N}_{1}+{N}_{2}-1} \gamma^{{N}_{1}+{N}_{2}-i} ||\bm{D}_{m}^{i-{N}_{1}}-\bm{d}_{gt}||_1\\
\end{aligned}
\end{equation}
where $\gamma=0.9$, and ${\bm{d}}_{gt}$ is the ground truth.

\begin{figure}[t]
    \centering
    \includegraphics[width=0.98\linewidth]{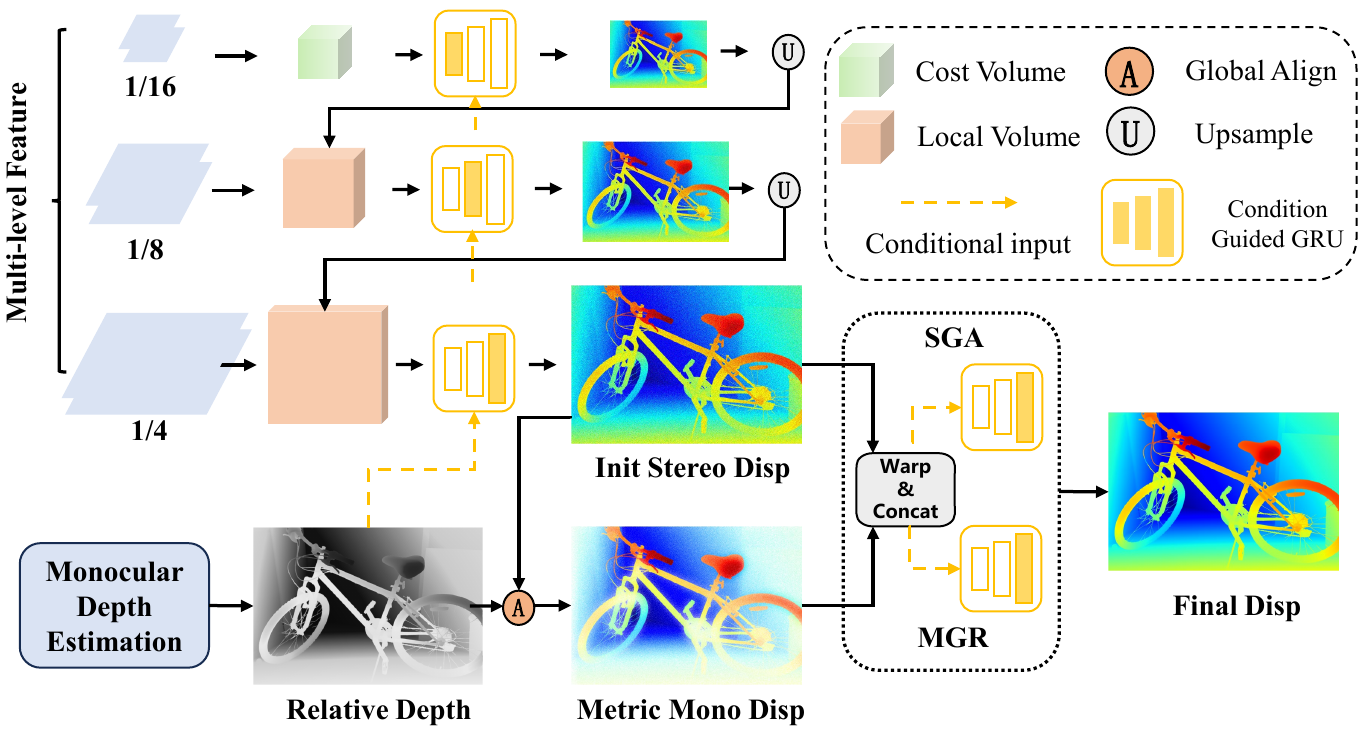}
    \caption{Overview of RT-MonSter++, the real-time variant of MonSter++, capable of achieving over 20 FPS inference at 1K resolution.}
    \label{fig:network_small}
\end{figure}

\subsection{Multi-view stereo}\label{sec:mvs}

The fundamental difference between multi-view stereo and stereo matching lies in their input: a set of unrectified images with arbitrary camera poses, where matching must be performed along more general epipolar geometries. This difference leads to a key distinction in the cost volume construction stage. 

Stereo matching only requires cost computation along horizontal directions between two views, we adopt the Geometry Encoding Volume following IGEV~\cite{igev} in our stereo model, as described in Sec. \ref{sec:2branch}. In multi-view stereo setting, we follow MVSNet~\cite{yao2018mvsnet} to construct a variance-based cost metric that encodes feature similarity across multiple unrectified views to replace the Geometry Encoding Volume.
For simplicity, in the following we denote $\mathbf{I}_1$ as the reference image, $\{\mathbf{I}_i\}_{i=2}^{N}$ the source images, and $\{\mathbf{K}_i, \mathbf{R}_i, \mathbf{t}_i\}_{i=1}^{N}$ the camera intrinsics, rotations and translations that correspond to the feature maps. Following MVSNet~\cite{yao2018mvsnet}, we wrap features from source views to the reference view using differentiable homography:
	\begin{equation}
	\mathbf{H}_i(d) = \mathbf{K}_i \cdot \mathbf{R}_i \cdot \Big(\mathbf{I} - \frac{(\mathbf{t}_1 - \mathbf{t}_i) \cdot \mathbf{n}_1^T}{d} \Big) \cdot \mathbf{R}_1^T \cdot \mathbf{K}_1^T.   
	\end{equation}
where $d \in \mathbb{B}$, $\mathbb{B}$ denotes the bins uniformly sampled in the log space from $d_{min}$ to $d_{max}$, which represents the depth search range. Then we use variance-based cost metric $\mathcal{M}$ for \textit{N}-view similarity measurement. We aggregate multiple feature volumes $\{\mathbf{V}_i\}_{i=1}^N$ to one cost volume $\mathbf{C}$ as:
	\begin{equation} \label{eq:metric}
	\mathbf{C} = \mathcal{M}(\mathbf{V}_1, \cdots, \mathbf{V}_N) = \frac{\sum\limits_{i=1}^N{(\mathbf{V}_i - {\mathbf{V}_i})^2}}{N}
	\end{equation}
The variance-based cost volume $\mathbf{C}$ shares the same dimensions as the Geometry Encoding Volume used in the stereo matching model, and is entirely parameter-free. The subsequent processing stages remain identical to those in the stereo matching pipeline.

\subsection{Real-time version of MonSter++}\label{sec:RT_monster}
Building upon MonSter++, we develop a real-time variant capable of achieving over 20 FPS inference speed at 1K resolution with only 2 GB of GPU memory. While MonSter++ performs cost aggregation and iterative refinement solely at the $1/4$ resolution level, our real-time model adopts a fully coarse-to-fine framework, progressively operating at $1/16$, $1/8$, and $1/4$ resolutions as shown in Fig.\ref{fig:network_small}. This hierarchical design enables early-stage pruning of the disparity search space, significantly improving efficiency. To further reduce computational overhead at higher resolutions ($1/8$ and $1/4$), we construct local cost volumes instead of full ones. This cascaded search and multi-scale depth fusion strategy enables us to fully exploit monocular priors while achieving both high efficiency and strong robustness. Specifically, for each pixel, we center the disparity sampling around the estimate from the previous resolution stage. Let $d_{\text{prev}}(i)$ denote the predicted disparity in pixel $i$ from the preceding stage, and $W$ denote the feature width at the current stage. Given the number of disparity samples $D$ and the sampling interval $\delta$, the local disparity range is defined as follows:
\begin{equation}
\begin{aligned}
d_{\min}(i) &= \max\left(0,\, d_{\text{prev}}(i) - \tfrac{D-1}{2}\cdot \delta\right), \\
d_{\max}(i) &= \min\left(W,\, d_{\text{prev}}(i) + \tfrac{D-1}{2}\cdot \delta\right).
\end{aligned}
\end{equation}
To account for boundary clipping, we define the actual sampling interval $\Delta$ as:
\begin{equation}
\Delta(i) = \frac{d_{\max}(i) - d_{\min}(i)}{D - 1}.
\end{equation}
Finally, the set of sampled disparities at that pixel is given by:
\begin{equation}
d_{\text{sample}}(i) = \left\{ d_{\min}(i) + n \cdot \Delta(i) \mid n = 0, 1, \ldots, D-1 \right\}.
\end{equation}
This local sampling strategy effectively narrows the disparity search range at each level, leading to a substantial speed-up without compromising accuracy.

To improve runtime efficiency while maintaining competitive accuracy, we make several adjustments to the model architecture. First, our real-time lightweight variant adopts a single-layer ConvGRU. Since our design follows a coarse-to-fine framework, only the hidden state corresponding to the current resolution stage needs to be updated, which further improves efficiency without sacrificing accuracy. Second, we reduce the number of GRU updates: one update at each of the two lower resolutions and two updates at 1/4 resolution. Finally, we redesign the cost aggregation module using a lightweight architecture to better balance speed and performance.

\section{Experiment}
\label{sec:experiment}

\begin{table*}[ht]
	\centering
	\caption{\textbf{Quantitative evaluation on Scene Flow test set.} The \textbf{best} result is bolded, and the \underline{second-best} result is underscored.}
	\begin{tabular}{c|ccccc|c}
		\toprule
		Method & 
		GwcNet\cite{gwcnet} & LEAStereo\cite{leastereo} & ACVNet\cite{acvnet} &IGEV\cite{igev} & Selective-IGEV\cite{selective-igev} & MonSter++ (Ours) \\ \midrule
		EPE (px)$\downarrow$
		& 0.76 
		& 0.78 
		& 0.48
		& 0.47 
		& \underline{0.44} 
		& \textbf{0.37} $_{\color[rgb]{1,0,0}(-15.91\%)}$  \\
		\bottomrule
	\end{tabular}

	\label{tab:sceneflow}
\end{table*}

\begin{table*}
\setlength{\tabcolsep}{4.5pt}
    \centering
    \caption{\textbf{Results on four popular stereo matching benchmarks.} All results are derived from official leaderboard publications or corresponding papers. All metrics are presented in percentages, except for RMSE, which is reported in pixels. For testing masks, “All” denotes being tested with all pixels while “Noc” denotes being tested with a non-occlusion mask. The \colorbox{red!25}{\textbf{best}} and \colorbox{yellow!25}{second best} are marked with colors.}
    \begin{tabular}{l|ccc|ccc|cccc|cccc}
     \toprule
     \multirow{3}{*}{Method} & \multicolumn{3}{c|}{ETH3D\cite{eth3d}} & \multicolumn{3}{c|}{Middlebury\cite{middlebury}} & \multicolumn{4}{c|}{KITTI 2015\cite{kitti2012}} & \multicolumn{4}{c}{ KITTI 2012\cite{kitti2012}}  \\
     \cline{2-15}
     & Bad1.0 & Bad1.0 & RMSE & Bad4.0 & RMSE & RMSE & D1-bg & D1-all & D1-bg & D1-all & Out-2 & Out-2 & Out-3 & Out-3\\
     & Noc & All & Noc & Noc & Noc & All & Noc  & Noc & All &  All & Noc & All & Noc & All \\
     
    \hline

    GwcNet\cite{gwcnet} & 6.42 & 6.95 & 0.69 & - & - & - & 1.61 & 1.92 & 1.74 & 2.11 & 2.16 & 2.71 & 1.32 & 1.70 \\
    GANet\cite{ganet}  & 6.22 & 6.86 & 0.75 & - & - & - & 1.40 & 1.73 & 1.55 & 1.93  & 1.89 & 2.50 & 1.19 & 1.60 \\
    LEAStereo\cite{leastereo}  & - & - & - & 2.75 & 8.11 & 13.7 & 1.29 & 1.51 & 1.40 & 1.65  & 1.90 & 2.39 & 1.13 & 1.45 \\
    ACVNet\cite{acvnet} & 2.58 & 2.86 & 0.45 & 7.99 & 32.2 & 38.6 & 1.26 & 1.52 & 1.37 & 1.65  & 1.83 & 2.35 & 1.13 & 1.47 \\
    RAFT-Stereo\cite{raft-stereo} & 2.44 & 2.60 & 0.36 & 2.75 & 8.41 & 12.6 & 1.44 & 1.69 & 1.58 & 1.82  & 1.92 & 2.42 & 1.30 & 1.66 \\
    CREStereo\cite{crestereo} & 0.98 & 1.09 & 0.28 & 2.04 & 7.70 & 10.5 & 1.33 & 1.54 & 1.45 & 1.69  & 1.72 & 2.18 & 1.14 & 1.46 \\
    IGEV\cite{igev} & 1.12 & 1.51 & 0.34 & 3.33 & 12.80 & 15.1 & 1.27 & 1.49 & 1.38 & 1.59  & 1.71 & 2.17 & 1.12 & 1.44 \\
    CroCo-Stereo\cite{croco} & 0.99 & 1.14 & 0.30 & 4.18 & 8.91 & 10.6 & 1.30 & 1.51 & 1.38 & 1.59  & - & - & - & - \\
    DLNR\cite{dlnr} & - & - & - & 1.89 & 7.78 & 10.2 & 1.42 & 1.61 & 1.60 & 1.76  & - & - & - & - \\
    Selective-IGEV\cite{selective-igev} & 1.23 & 1.56 & 0.29 & 1.36 & 7.26 & 9.26 & 1.22 & 1.44 & 1.33 & 1.55  & 1.59 & 2.05 & 1.07 & 1.38 \\
    LoS\cite{los} & 0.91 & 1.03 & 0.31 & 2.30 & 6.99 & 8.78 & 1.29 & 1.52 & 1.42 & 1.65  & 1.69 & 2.12 & 1.10 & 1.38 \\
    NMRF-Stereo\cite{nmrf} & - & - & - & - & - & - &\cellcolor{yellow!25}1.18 & 1.46 & \cellcolor{yellow!25}1.28 & 1.57  & 1.59 & 2.07 & \cellcolor{yellow!25}1.01 & \cellcolor{yellow!25}1.35 \\
    FoundationStereo\cite{foundationstereo} & \cellcolor{yellow!25}0.26 & \cellcolor{yellow!25}0.48 & \cellcolor{yellow!25}0.20 & \cellcolor{red!25}\textbf{1.04} & \cellcolor{yellow!25}6.48 & \cellcolor{yellow!25}8.39 & - & - & - & -  & - & - & - & - \\
    IGEV++~\cite{igev++} & 1.14 & 1.58 & 0.74 & 1.82 & 7.23 & 10.20 & 1.20 & \cellcolor{yellow!25}1.42 & 1.31 & \cellcolor{yellow!25}1.51 & \cellcolor{yellow!25}1.56 & \cellcolor{yellow!25}2.03 & 1.04 & 1.36 \\
    MonSter++ (Ours) &\cellcolor{red!25}\textbf{0.25} & \cellcolor{red!25}\textbf{0.45} &\cellcolor{red!25}\textbf{0.18} & \cellcolor{yellow!25}1.18 & \cellcolor{red!25}\textbf{6.03} & \cellcolor{red!25}\textbf{7.81} & \cellcolor{red!25}\textbf{1.02} & \cellcolor{red!25}\textbf{1.29} & \cellcolor{red!25}\textbf{1.12} &\cellcolor{red!25}\textbf{1.37}  & \cellcolor{red!25}\textbf{1.30} & \cellcolor{red!25}\textbf{1.70} & \cellcolor{red!25}\textbf{0.79} & \cellcolor{red!25}\textbf{1.07} \\
    \bottomrule
    \end{tabular}
\label{tab:benchmark}
\end{table*}

\subsection{Implementation}
We implement MonSter++ with PyTorch and perform experiments using NVIDIA RTX 4090 GPUs. We use the AdamW \cite{adamw} optimizer and clip gradients to the range [-1, 1] following baseline~\cite{igev}. We use the one-cycle learning rate schedule with a learning rate of 2e-4 and train MonSter++ with a batch size of 8 for 200k steps as the pretrained model. For the monocular branch, we use the ViT-large version of DepthAnythingV2~\cite{depthany_v2} and freeze its parameters to prevent training of stereo-matching tasks from affecting its generalization and accuracy~\footnote{All research undertaken at Meta AI was limited to general guidance on model architectural design.  Meta did not participate in any model training activities.}.

Following the standard~\cite{crestereo, los, selective-igev, foundationstereo}, we adopt a two-stage training strategy for the evaluation on the ETH3D and Middlebury benchmarks. In the first training stage, we follow SOTA methods~\cite{crestereo, los, selective-igev, foundationstereo} to create the Basic Training Set (BTS) from various public datasets for pretraining, including Scene Flow~\cite{dispNetC}, FSD~\cite{foundationstereo}, CREStereo~\cite{crestereo}, Tartan Air~\cite{tartanair}, Sintel Stereo~\cite{sintel}, FallingThings \cite{falling}, and InStereo2k~\cite{instereo2k}. To develop a highly generalizable stereo model for open-source community use, we additionally collected and aggregated a wide range of public datasets, resulting in over 2 million image pairs for training. We refer to this combined dataset as the Full Training Set (FTS). Compared to the Basic Training Set (BTS), the Full Training Set (FTS) additionally includes data from 3DKenBurns~\cite{3dkenburns}, DynamicStereo~\cite{dynamicstereo}, IRS~\cite{irs}, VA~\cite{VA}, Booster~\cite{booster}, and Carla-Highres~\cite{carlahierarchical} datasets.

\subsection{Stereo matching Benchmark Performance}
\label{sec:4.2}
To demonstrate the outstanding performance of our method, we independently evaluate MonSter++ on both stereo matching and multi-view stereo tasks. We first evaluate our method on the stereo matching task using the five most commonly used benchmarks: KITTI 2012 \cite{kitti2012}, KITTI 2015 \cite{kitti2015}, ETH3D \cite{eth3d}, Middlebury \cite{middlebury}, and Scene Flow~\cite{dispNetC}.

\textbf{Scene Flow}~\cite{dispNetC}. As shown in Tab. \ref{tab:sceneflow}, we achieve a new state-of-the-art performance with an EPE metric of 0.37 on Scene Flow, surpassing our baseline~\cite{igev} by 21.28\% and outperforming the SOTA method~\cite{selective-igev} by 15.91\%.

\textbf{ETH3D}\cite{eth3d}. Following the SOTA methods \cite{crestereo, los, igev++, unifying, selective-igev}, the full training set is composed of the BTS and ETH3D training set.

\begin{figure*}[ht]
    \centering
    \includegraphics[width=0.95\linewidth]{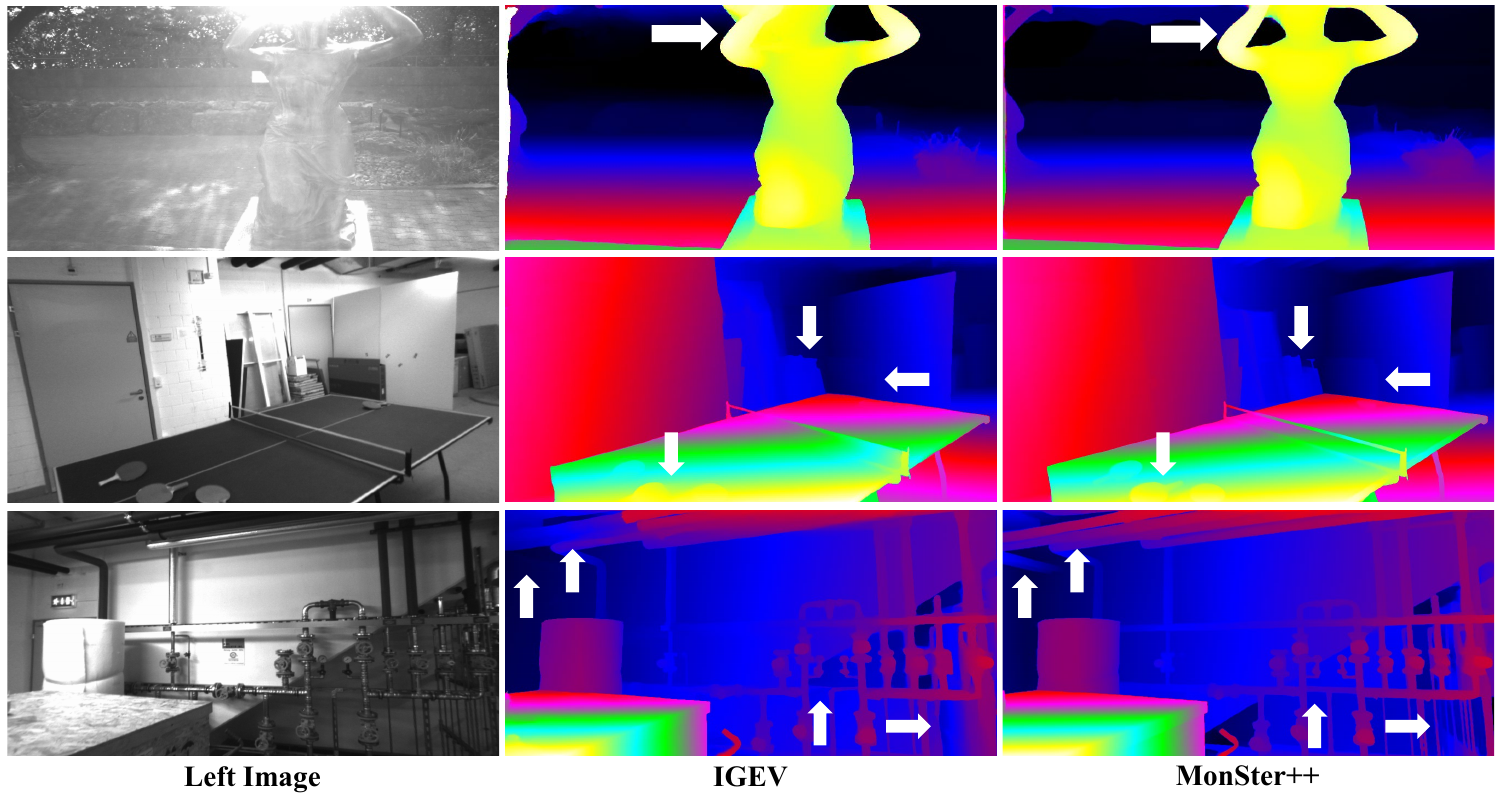}
    \caption{\textbf{Qualitative results on ETH3D.} MonSter++ outperforms IGEV in challenging areas with strong reflectance, fine structures etc.}
    \label{fig:vis_eth}
\end{figure*}

Our MonSter++ ranks 1st on the ETH3D leaderboard. As shown in Tab. \ref{tab:benchmark}, MonSter++ achieves the best performance among all published methods. Compared with baseline IGEV, we achieve improvements of 77.68\%, 70.20\%, and 47.06\% in the three reported metrics, respectively. Qualitative comparisons shown in Fig.\ref{fig:vis_eth} exhibit a similar trend. Notably, even compared with the previous best method FoundationStereo, we improved the RMSE (Noc) metric from 0.20 to 0.18, achieving a 10\% improvement. 

\begin{table*}[ht]
  \centering
  \caption{\textbf{Results on two popular multi-view stereo depth estimation benchmarks.} Note that the $*$ marks the result reproduced by us using their open-source code, other reported numbers are from ~\cite{afnet} and the corresponding original papers. The \colorbox{red!25}{\textbf{best}} and \colorbox{yellow!25}{second best} are marked with colors.}
  \begin{tabular}{c|ccc|ccccc}
  \toprule
    \multirow{2}{*}{Method} & \multicolumn{3}{c|}{DDAD~\cite{ddad}}  & \multicolumn{5}{c}{KITTI Eigen Split~\cite{kitti_eigen}} \\ 
    \cline{2-9}
     & AbsRel$\downarrow$  & SqRel$\downarrow$  & RMSE$\downarrow$  & AbsRel$\downarrow$  & SqRel$\downarrow$  & RMSE$\downarrow$ & {RMSE}$_{log}$$\downarrow$ & $\delta<$ 1.25 $\uparrow$ \\
    \hline
    Deepv2d~\cite{teed2018deepv2d} & - & - & - & 0.091 & 0.582 & 3.644 & 0.154 & 0.923\\
    CasMVS~\cite{gu2020cascade} & 0.129  & 2.01  & 9.87  & 0.066  & 0.228 & 2.567  & 0.112 & 0.945\\
    MVSNet~\cite{yao2018mvsnet}  & 0.109  & 1.62  & 8.21  & - & - & - & - & -\\
    IterMVS~\cite{itermvs}  & 0.104  & 1.59  & 7.95  & 0.057  & 0.178  & 2.234 & 0.087 & 0.968 \\
    MVS2D~\cite{mvs2d}  & 0.132  & 2.05  & 9.82  & 0.058  & 0.176  & 2.277 & 0.090 & 0.966 \\
    SC-GAN~\cite{wu2019spatial} & - & - & - & 0.063 &  0.178 & 2.129 & 0.097 & 0.961 \\
    MaGNet~\cite{bae2022multi}  & 0.112  & 1.74  & 9.23  & 0.054 & 0.162 & 2.158 & 0.083 & 0.971 \\
  AFNet~\cite{afnet} & \cellcolor{yellow!25}0.088 & \cellcolor{yellow!25}1.41 & \cellcolor{yellow!25}7.23 &  0.044 &  0.132 &  1.712 & 0.069 & 0.980\\
  MVSFormer++~\cite{mvsformer++}  & $0.092^*$  & $1.52^*$  & $7.56^*$  &  \cellcolor{yellow!25}$0.041^*$ &  \cellcolor{yellow!25}$0.127^*$ & \cellcolor{yellow!25}$1.649^*$ & \cellcolor{yellow!25}$0.067^*$ & \cellcolor{yellow!25}$0.983^*$\\
  MonSter++ (Ours) & \cellcolor{red!25}\textbf{0.075} & \cellcolor{red!25}\textbf{1.21} &  \cellcolor{red!25}\textbf{6.45} &  \cellcolor{red!25}\textbf{0.037} &  \cellcolor{red!25}\textbf{0.104} &  \cellcolor{red!25}\textbf{1.526} & \cellcolor{red!25}\textbf{0.060} & \cellcolor{red!25}\textbf{0.987}\\
    \bottomrule
  \end{tabular}
  \label{tab:mvs}
\end{table*}

\textbf{Middlebury.} Also following ~\cite{crestereo,los, unifying, selective-igev}, the training set of Middlebury is the combination of the BTS and Middlebury training set. As shown in Tab. \ref{tab:benchmark}, MonSter++ outperforms all existing methods in terms of the RMSE metric. Compared with baseline \cite{igev}, we achieve an improvement of 64.56\% on the Bad 4.0 (Noc) metric. Compared with the SOTA method FoundationStereo \cite{foundationstereo}, we achieve a 6.94\% improvement in the RMSE (Noc) metric.

\textbf{KITTI.} Following the training of SOTA methods ~\cite{crestereo, nmrf, selective-igev}, we finetuned the Scene Flow pretrained model on the mixed dataset of KITTI 2012 and KITTI 2015 for 50k steps. At the time of writing, MonSter++ ranks $1^{st}$  simultaneously on both the KITTI 2012 and KITTI 2015 leaderboards. As shown in Tab. \ref{tab:benchmark}, we achieve the best performance for all metrics. On KITTI 2015, MonSter++ surpasses CREStereo \cite{crestereo} and Selective-IGEV~\cite{selective-igev} by 18.93\% and 11.61\% on the D1-all metric of all regions, respectively. As for KITTI 2012, we significantly outperform the existing SOTA by a large margin. Compared to the current highest-precision methods, IGEV++~\cite{igev++} and NMRF-Stereo \cite{nmrf}, we achieve a 16.26\% improvement and a 20.74\% improvement respectively in the Out-2 metric and the Out-3 metric across all regions.

\subsection{Multi-view Stereo Benchmark Performance}
\label{sec:4.3}
To demonstrate the outstanding performance of our method in the multi-view stereo task, we evaluate MonSter++ on the two popular multi-view stereo depth estimation benchmarks: DDAD~\cite{ddad} and KITTI Eigen split~\cite{kitti_eigen}.

\textbf{DDAD}~\cite{ddad}. We employ the same training and testing scheme to all methods, following~\cite{afnet}. For all methods, we train from scratch on DDAD training set. During testing, we evaluate all ring cameras to report the average performance, instead of using only the front-view camera. As shown in Tab. \ref{tab:mvs}, our MonSter++ achieves the state-of-the-art (SOTA) performance on DDAD. Compared with the previous best method~\cite{afnet},  MonSter++ achieves over 14.8\% improvement on AbsRel error.

\textbf{KITTI}~\cite{kitti_eigen}. KITTI Eigen split~\cite{kitti_eigen} is a classical benchmark for multi-view depth estimation. Thus, we compare our method with state-of-the-art methods on this dataset. As shown in Tab. \ref{tab:mvs}, MonSter++ outperforms all existing methods across all evaluation metrics. It is worth mentioning that MonSter++ achieves a 18.11\% improvement and a 21.21\% improvement respectively on SqRel error compared with the current highest-precision methods~\cite{mvsformer++} and~\cite{afnet}.

\subsection{Performance in Ill-posed Regions}
MonSter++ fully leverages the advantages of monocular depth priors, effectively overcoming challenges in ill-posed regions. To validate this, we conduct comparisons on several representative ill-posed regions, such as reflective areas, edge and non-edge regions, and distant backgrounds:

\begin{table*}
    \centering
    \caption{Quantitative comparisons on the KITTI 2012~\cite{kitti2012} and KITTI 2015~\cite{kitti2015} stereo matching benchmarks. Our RT-MonSter++ runs 4 update iterations at inference. The \colorbox{red!25}{\textbf{best}} and \colorbox{yellow!25}{second best} are marked with colors.}
    \begin{tabular}{l|cccccccc|cccc|c}
     \toprule
     \multirow{3}{*}{Method} & \multicolumn{8}{c|}{KITTI 2012~\cite{kitti2012}} & \multicolumn{4}{c|}{ KITTI 2015~\cite{kitti2015}} & \multirow{3}{*}{\makecell{Run-time\\(ms)}} \\
     \cline{2-13}
      & Out-2 & Out-2 & Out-3 & Out-3 & Out-4 & Out-4 & Out-5 & Out-5 & D1-fg & D1-all & D1-fg & D1-all & \\
       & Noc & All & Noc & All & Noc & All & Noc & All & Noc  & Noc & All & All &\\
      \cline{1-14}

     HITNet~\cite{hitnet} & 2.00 & 2.65 & 1.41 & 1.89 & 1.14 & 1.53 & 0.96 & 1.29 & \cellcolor{yellow!25}2.72 & 1.74 & \cellcolor{yellow!25}3.20 & 1.98 & \textbf{20}\\
    DeepPrunerFast~\cite{deeppruner} & - & - & - & - & - & - & - & - & 3.43 & 2.35 & 3.91 & 2.59 & 50\\
     AANet~\cite{aanet} & 2.30 & 2.96 & 1.55 & 2.04 & 1.20 & 1.58 & 0.98 & 1.30 &  3.66 & 1.85 & 3.96 & 2.03 & 60\\
    BGNet+~\cite{bgnet}  & 2.78 & 3.35 & 1.62 & 2.03 & 1.16 & 1.48 & 0.90 & 1.16 & 3.76 & 2.01 & 4.09 & 2.19 & 35\\
    CoEx~\cite{coex} & 2.54 & 3.09 & 1.55 & 1.93 &  1.15 & 1.42 & 0.91 & 1.13 & 3.09 & 1.86 & 3.41 & 2.02 & 27\\
    Fast-ACVNet+~\cite{fast-acv} & 2.39 & 2.97 & 1.45 & 1.85 & 1.06 & 1.36 & 0.85 & 1.09 & 3.29 & 1.85 & 3.53  & 2.01 & 45\\
    RT-IGEV++~\cite{igev++} & \cellcolor{yellow!25}1.93 & \cellcolor{yellow!25}2.51 & \cellcolor{yellow!25}1.29 & \cellcolor{yellow!25}1.68 & \cellcolor{yellow!25}1.00 & \cellcolor{yellow!25}1.30 & \cellcolor{yellow!25}0.81 & \cellcolor{yellow!25}1.06  & 3.17 & \cellcolor{yellow!25}1.64 & 3.37 & \cellcolor{yellow!25}1.79 & 48 \\
    RT-MonSter++ (Ours) & \cellcolor{red!25}\textbf{1.75} & \cellcolor{red!25}\textbf{2.26} & \cellcolor{red!25}\textbf{1.07} & \cellcolor{red!25}\textbf{1.41} & \cellcolor{red!25}\textbf{0.80} & \cellcolor{red!25}\textbf{1.05} & \cellcolor{red!25}\textbf{0.65} & \cellcolor{red!25}\textbf{0.84}  & \cellcolor{red!25}\textbf{2.49} & \cellcolor{red!25}\textbf{1.52} & \cellcolor{red!25}\textbf{2.78} & \cellcolor{red!25}\textbf{1.69} & 47 \\
    \bottomrule
    \end{tabular}
\label{tab:kitti_realtime}
\end{table*}

\begin{table}[t]
\setlength{\tabcolsep}{4.5pt}
    \centering
    \caption{\textbf{Results of the reflective regions on KITTI 2012 leaderboard.} The \colorbox{red!25}{\textbf{best}} and \colorbox{yellow!25}{second best} are marked with colors.}
    \begin{tabular}{l|cccccc}
     \toprule
     \multirow{3}{*}{Method} & \multicolumn{6}{c}{ KITTI 2012 Reflective Region} \\
     \cline{2-7}
     & Out-2 & Out-2 & Out-3 & Out-3 & Out-4 & Out-4\\
     & Noc & All & Noc & All & Noc & All  \\
     
    \hline

    ACVNet\cite{acvnet} & 11.42 & 13.53 & 7.03 & 8.67 & 5.18 & 6.48\\
    CREStereo\cite{crestereo} & 9.71 & 11.26 & 6.27 & 7.27 & 4.93 & 5.55\\
    IGEV\cite{igev} & 7.57 & 8.80 & 4.35 & 5.00 & 3.16 & 3.57 \\
    Selective-IGEV\cite{selective-igev} & 6.73 & \cellcolor{yellow!25}7.84 & 3.79 & \cellcolor{yellow!25}4.38 & 2.66 & 3.05  \\
    LoS\cite{los} & \cellcolor{yellow!25}6.31 & \cellcolor{yellow!25}7.84 & \cellcolor{yellow!25}3.47 & 4.45 & \cellcolor{yellow!25}2.41 & \cellcolor{yellow!25}3.01 \\
    NMRF-Stereo\cite{nmrf} & 10.02 & 12.34 & 6.35 & 8.11 & 4.80 & 6.09\\
    MonSter++ (Ours) & \cellcolor{red!25}\textbf{5.68} & \cellcolor{red!25}\textbf{6.66} & \cellcolor{red!25}\textbf{2.91} & \cellcolor{red!25}\textbf{3.45} & \cellcolor{red!25}\textbf{1.92} & \cellcolor{red!25}\textbf{2.29}\\
    \bottomrule
    \end{tabular}
\label{tab:reflective}
\end{table}

\begin{table}[t]
\centering
\caption{\textbf{Comparison on SceneFlow test set in different regions.} The \textbf{best} result is bolded.}
\begin{tabular}{lcccc}
\midrule
\multirow{2}{*}{Method} & \multicolumn{2}{c}{Edges}  & \multicolumn{2}{c}{Non-Edges}\\ \cline{2-5} & EPE & \textgreater{}1px & EPE &\textgreater{}1px\\ \hline
IGEV~\cite{igev} & 2.23 & 20.42 & 0.41 & 4.58 \\
Selective-IGEV~\cite{selective-igev}& 2.18 & 20.01  & 0.38 & 4.35  \\ 
MonSter++ (Ours) & \textbf{1.91} & \textbf{18.59} & \textbf{0.31} & \textbf{3.57}\\
\midrule
\end{tabular}

\label{tab:edge}
\end{table}

\begin{itemize}
    \item \textbf{Reflective Regions}: We conducted comparisons on the reflective regions of KITTI 2012 benchmark. MonSter++ ranks $1^{st}$ on the KITTI 2012 leaderboard for all metrics of reflective regions. As shown in Tab.\ref{tab:reflective}, we have elevated the existing SOTA to a new level. MonSter++ surpasses Selective-IGEV and LoS by 24.92\% and 23.92\% on Out-4 (All) metric, respectively. Notably, compared to the SOTA method NMRF-Stereo, we achieved significant improvements of 57.46\% and 62.40\% on the Out-3 (All) and Out-4 (All) metrics respectively.
    
    \item \textbf{Edge$\And$Non-edge Regions}: Stereo matching faces challenges in edge and low-texture regions, which our method addresses by leveraging the strengths of monocular depth. To evaluate MonSter++ in these areas, we divide the Scene Flow test set into edge and non-edge regions using the Canny operator following \cite{selective-igev}. As shown in Tab.~\ref{tab:edge}, MonSter++ outperforms the baseline~\cite{igev} by 14.35\% and 24.39\% in edge regions and non-edge regions. Even when compared to \cite{selective-igev}, which is specifically designed to address edge and textureless regions, our method achieves improvements of 12.39\% and 18.42\% in two types of regions on EPE metric, respectively.
    
    \item \textbf{Distant Backgrounds}: Stereo matching struggles with depth perception for distant objects, and we improve it by incorporating our SGA and MGR modules. As shown in Tab.\ref{tab:benchmark}, MonSter++ improves the D1-bg metric by 18.84\% compared to the baseline on KITTI 2015 benchmark. The D1-bg metric reflects the percentage of stereo disparity outliers averaged specifically over background regions.
\end{itemize}

\begin{table*}[t] 
  \centering 
  \caption{ Zero-shot generalization benchmark for real-time stereo matching methods and High-accuracy stereo matching methods. The \colorbox{red!25}{\textbf{best}} and \colorbox{yellow!25}{second best} are marked with colors.} 
  \label{tab:generalization} 
  \begin{tabular}{l|lcccc|c|c} 
    \toprule 
    \multirow{2}{*}{Target} & \multirow{2}{*}{Method} & KITTI-12 & KITTI-15 & ETH3D & Middlebury & Avg. & Runtime \\ 
    & & ($>$3px) & ($>$3px) & ($>$1px) & ($>$2px) & & (ms) \\ 
    \hline 
    \multirow{7}{*}{ \rotatebox{90}{\textit{Real-Time}}} & \textbf{Training Set} & \multicolumn{4}{c|}{\textbf{Scene Flow}} &\multicolumn{1}{c|}{} \\ 
    \cline{2-8} 
    & DeepPrunerFast~\cite{deeppruner} & 16.8 & 15.9 & 11.0 & 30.8 & 18.63 & 50\\ 
    & BGNet+~\cite{bgnet} & 24.8 & 20.1 & 10.3 & 37.0 & 23.05 & 35 \\ 
    & CoEx~\cite{coex} & 13.5 & 11.6 & 9.0 & 25.5 & 14.90 & 27 \\ 
    & Fast-ACVNet+~\cite{fast-acv} & 12.4 & 10.6 & 7.9 & 20.1 & 12.75 & 45 \\ 
    & RT-IGEV++\cite{igev++} & \cellcolor{yellow!25}5.9 & \cellcolor{yellow!25}6.6 & \cellcolor{red!25}\textbf{5.8} & \cellcolor{yellow!25}17.3 & \cellcolor{yellow!25}8.90 & 48 \\ 
    & RT-MonSter++ (Ours)& \cellcolor{red!25}\textbf{4.8} & \cellcolor{red!25}\textbf{5.9} & \cellcolor{yellow!25}6.0 & \cellcolor{red!25}\textbf{11.4} & \cellcolor{red!25}\textbf{7.03} & 47\\ 
    \hline 
    \multirow{8}{*}{ \rotatebox{90}{\textit{Accuracy}}} & \textbf{Training Set} & \multicolumn{4}{c|}{\textbf{Scene Flow}} &\multicolumn{1}{c|}{}\\ 
    \cline{2-8} 
    & CFNet\cite{cfnet} & \cellcolor{yellow!25}4.7 & 5.8 & 5.8 & 9.8 & 6.53 & 180\\ 
    & RAFT-Stereo\cite{raft-stereo} & 5.1 & 5.7 & \cellcolor{yellow!25}3.3 & 9.4 & 5.88 & 380 \\ 
    & CREStereo\cite{crestereo} & 5.0 & 5.8 & 9.0 & 12.9 & 8.18 & 410\\ 
    & Selective-IGEV\cite{selective-igev} & 5.6 & 6.1 & 5.4 & 12.0 & 7.28 & 240 \\ 
    & IGEV\cite{igev} & 4.8 & \cellcolor{yellow!25}5.5 & 3.6 & \cellcolor{yellow!25}6.2 & \cellcolor{yellow!25}5.03 & 180\\ 
    & IGEV++\cite{igev++} &6.2 &6.4 &4.1 & 7.8 &6.13 & 280\\ 
    & MonSter++ (Ours)& \cellcolor{red!25}\textbf{3.6} & \cellcolor{red!25}\textbf{4.0} & \cellcolor{red!25}\textbf{2.0} & \cellcolor{red!25}\textbf{5.2} & \cellcolor{red!25}3.70 & 450\\ 
    \hline 
    \multirow{6}{*}{ \rotatebox{90}{\textit{Mix-Training}}} & \textbf{Training Set} & \multicolumn{4}{c|}{\textbf{ Full Training Set (FTS)}} &\multicolumn{1}{c|}{}\\ 
    \cline{2-8} 
    & RT-IGEV++\cite{igev++} & 3.9 & 4.6 & 2.5 & 8.3 & 4.83 & 48\\ 
    & RT-MonSter++ (Ours)& \cellcolor{yellow!25}3.2 & \cellcolor{yellow!25}3.7 & \cellcolor{yellow!25}1.7 & 6.2 & 3.70 & 47\\ 
    \cdashline{2-8} 
    \addlinespace[6pt] 
    & IGEV\cite{igev} & 3.7 & 3.9 & 2.0 & \cellcolor{yellow!25}4.3 & \cellcolor{yellow!25}3.48 & 180\\ 
    & MonSter++ (Ours)& \cellcolor{red!25}\textbf{2.8} & \cellcolor{red!25}\textbf{3.0} & \cellcolor{red!25}\textbf{0.8} & \cellcolor{red!25}\textbf{2.6} & \cellcolor{red!25}\textbf{2.30} & 450 \\ 
    \bottomrule 
  \end{tabular} 
\label{tab:zero-all} 
\end{table*}

\subsection{Real-Time Version of MonSter++}

To enable practical deployment on mobile platforms such as UAVs and robots, we design a real-time lightweight version of MonSter++ for the stereo matching task, named RT-MonSter++. Benefiting from MonSter++’s effective exploitation of monocular priors and a carefully designed coarse-to-fine strategy, RT-MonSter++ achieves over 20 FPS inference speed at 1K resolution while significantly surpassing the accuracy of state-of-the-art methods. Following IGEV++~\cite{igev++}, we conduct inference at the KITTI resolution of $1248\times384$ and compare our RT-MonSter++ with state-of-the-art real-time approaches on KITTI 2012 and KITTI 2015 benchmarks. As shown in Tab. \ref{tab:kitti_realtime}, RT-MonSter++ achieves the best performance among all published methods. On the KITTI 2012 benchmark, RT-MonSter++ surpasses the previous best method, RT-IGEV++~\cite{igev++}, by 19.23\% and 20.75\% on the Out-4 (ALL) and Out-5 (ALL) metrics, respectively. On the KITTI 2015 benchmark, RT-MonSter++ outperforms CoEx~\cite{coex} and RT-IGEV++~\cite{igev++} by 19.42\% and 21.45\% on D1-fg (Noc) metric, respectively.

Benefiting from the advantages of monocular priors, RT-MonSter++ also achieves the best generalization performance among real-time methods. We evaluate its generalization ability on the half-resolution training set of the Middlebury 2014 dataset~\cite{middlebury}, the ETH3D~\cite{eth3d} training set, and the training sets of KITTI 2012~\cite{kitti2012} and KITTI 2015~\cite{kitti2015}. As shown in Tab. \ref{tab:zero-all}, our RT-MonSter++ achieves state-of-the-art performance in the zero-shot setting among real-time stereo matching methods. Especially on the challenging Middlebury dataset with large disparities, RT-MonSter++ improves upon the previous best method, RT-IGEV++, by reducing the 2-pixel outlier rate from 17.3 to 11.4, corresponding to a 34.1\% improvement. 

It is worth highlighting that RT-MonSter++ demonstrates the strongest scalability among real-time methods. When trained on the Full Training Set (FTS), RT-MonSter++ achieves remarkable zero-shot performance gains over RT-IGEV++ across all four benchmarks, with particularly significant improvements of 32.0\% on ETH3D and 25.3\% on Middlebury. Moreover, RT-MonSter++ delivers performance comparable to the accuracy-oriented IGEV model, while requiring only one-third of its inference cost. We will release RT-MonSter++ model that trained on over 2 million image pairs to the community to empower this highly efficient and generalizable method for real-world deployment of depth perception on mobile platforms.

\begin{table}[t]
  \centering
  \caption{
   \textbf{Zero-shot generalization performance on DrivingStereo~\cite{drivingstereo} under different weather.} We utilize $>$3px All in comparisons. The \colorbox{red!25}{\textbf{best}} and \colorbox{yellow!25}{second best} are marked with colors.}
  \label{tab:generalization}
  \begin{tabular}{l|cccc|c}
    \toprule
    Method & Sunny & Cloudy & Rainy & Foggy & Avg.\\
    \hline    
    Nerf-Stereo\cite{nerfstereo} & \cellcolor{yellow!25}2.88 & 2.95 & 8.47 & 3.41 & 4.43 \\
    ZeroStereo\cite{zerostereo} & 3.15 & \cellcolor{yellow!25}2.69 & 11.71 & \cellcolor{red!25}\textbf{1.70} & 4.81 \\
    FoundationStereo\cite{foundationstereo} & 3.22 & 2.81 & 11.20 & \cellcolor{yellow!25}2.50 & 4.93 \\
    SMoEStereo\cite{SMoEStereo} & 3.51 & 3.11 & \cellcolor{yellow!25}6.08 & 4.77 & \cellcolor{yellow!25}4.37 \\
    MonSter++ (Ours)& \cellcolor{red!25}\textbf{2.60} & \cellcolor{red!25}\textbf{2.12} & \cellcolor{red!25}\textbf{3.08} & 2.94 & \cellcolor{red!25}\textbf{2.69} \\
    \bottomrule
  \end{tabular}
\label{tab:zero_driving}
\end{table}

\begin{figure*}[t]
    \centering
    \includegraphics[width=0.99\linewidth]{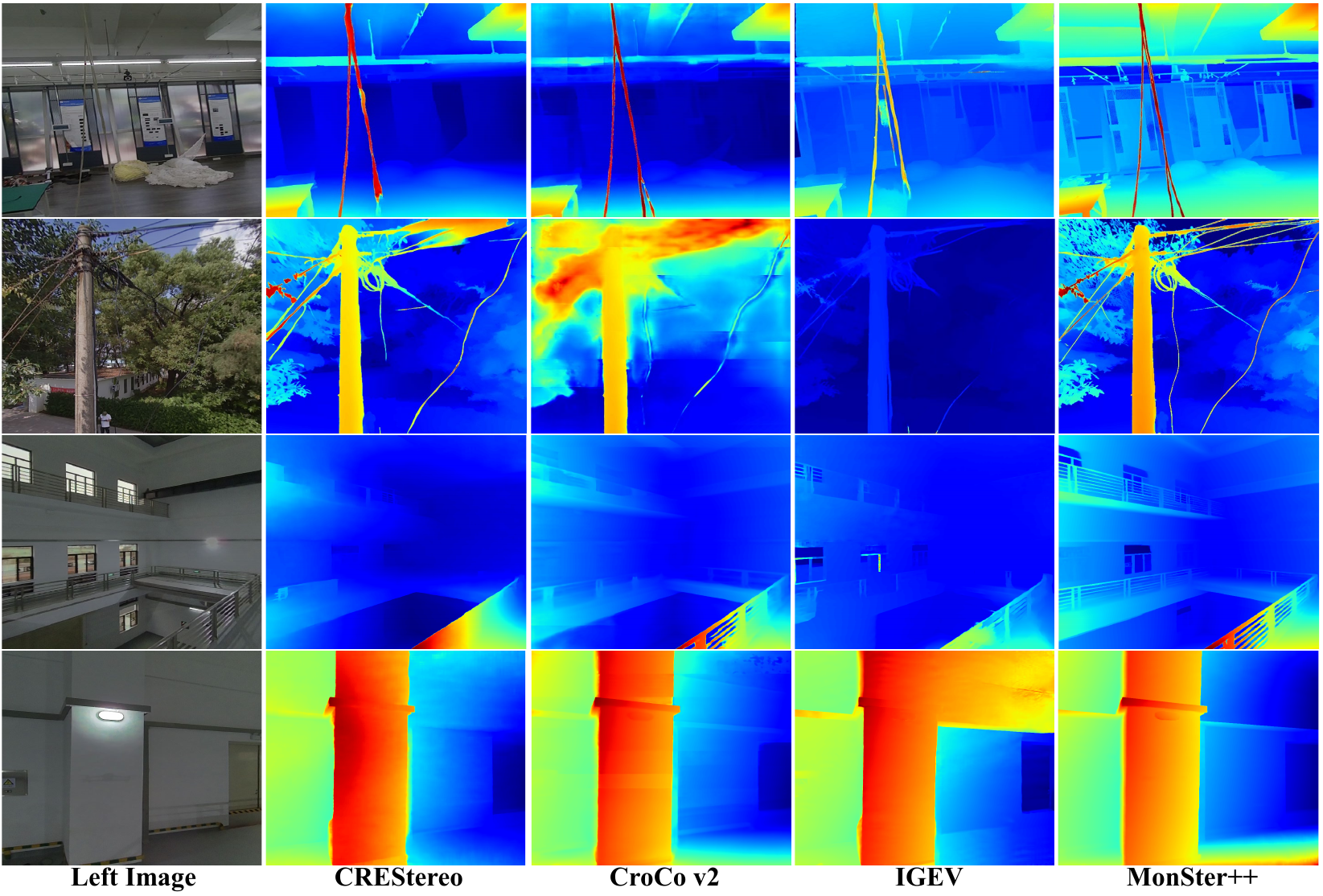}
    \caption{\textbf{Visual comparisons on our captured real-world zero-shot data provide a more comprehensive assessment of generalization capability.} All models are trained solely on the synthetic SceneFlow dataset. MonSter++ significantly outperforms IGEV in textureless regions, reflective areas and fine structures, etc.}
    \label{fig:vis_dt}
\end{figure*}

\subsection{Zero-shot Generalization}
Our method not only improves the accuracy of in-domain datasets but also enhances the generalization ability. We evaluate the generalization performance of MonSter++ from synthetic datasets to unseen real-world scenes. All models were trained on the same synthetic datasets and then tested directly on real-world datasets: KITTI\cite{kitti2012,kitti2015}, Middlebury\cite{middlebury}, and ETH3D\cite{eth3d} training sets. We conducted a more thorough generalization comparison by training with different datasets: 1) \textbf{Scene Flow}: We compared with SOTA methods which are only trained on the Scene Flow training set. As shown in Tab.~\ref{tab:zero-all}, MonSter++ achieves the best generalization performance across four datasets in accuracy models. Compared to our baseline method IGEV, we reduce the average error across four datasets from 5.03 to 3.70, achieving a 26.4\% improvement. MonSter++ even surpasses CFNet and CREStereo, both of which are specifically designed for cross-domain generalization. Fig.~\ref{fig:vis_kitti} highlights MonSter++’s improvements in reflective, textureless, and fine-structured regions. 2) \textbf{Full Training Set}: 
To demonstrate the scalability of our method, we further compare its generalization performance with the baseline method when trained on the Full Training Set.
As shown in the Mix Training setting in Tab. \ref{tab:zero-all}, our MonSter++ surpass the baseline method IGEV\cite{igev} by 39.5\% and 60.0\% on Middlebury and ETH3D, respectively. Notably, under the same mix-training setting, RT-MonSter++ delivers performance comparable to the accuracy-oriented IGEV model while requiring only 47 ms per inference—approximately one-third of IGEV’s runtime (180 ms).

\textbf{DrivingStereo}: In addition, we compare our method with state-of-the-art approaches specifically designed for generalization on the DrivingStereo\cite{drivingstereo} dataset. We choose this dataset for evaluation because existing generalization methods often rely on heterogeneous training data, such as incorporating private datasets or synthesizing right-view images, which makes direct comparison less fair. In contrast, the DrivingStereo dataset is unseen during training for all methods and contains diverse real-world scenarios with challenging weather conditions, providing a fair benchmark to assess a model’s ability to generalize to the real world. As shown in Tab. \ref{tab:zero_driving}, MonSter++ achieves the best overall performance across all weather conditions. In particular, on rainy scenes, it reduces the 3px error rate from 6.08 to 3.08 compared with the previous best method SMoEStereo \cite{SMoEStereo}, yielding a remarkable 49.3\% improvement. Furthermore, compared with the state-of-the-art FoundationStereo, MonSter++ lowers the average error across four weather conditions from 4.93 to 2.69, demonstrating its outstanding generalization ability in challenging environments.


\begin{table*}[t]
\centering
\caption{\textbf{Ablation study of the effectiveness of proposed modules on the Scene Flow test set.} All runtime results are measured under the same setting of 32 iterations. The \textbf{best} result is bolded.}
\begin{tabular}{l|cccc|cc|c}
\hline
Model                      & \begin{tabular}[c]{@{}c@{}}Monocular \\ Depth\end{tabular}  & \begin{tabular}[c]{@{}c@{}}Disparity \\ Fusion\end{tabular}   & \begin{tabular}[c]{@{}c@{}}Scale$\And$Shift \\ Refinement\end{tabular}  & \begin{tabular}[c]{@{}c@{}}Feature \\ Sharing\end{tabular}   & \begin{tabular}[c]{@{}c@{}}EPE\\ (px)\end{tabular} & \begin{tabular}[c]{@{}c@{}}\textgreater{}1px\\ (\%)\end{tabular} & \begin{tabular}[c]{@{}c@{}}Run-time\\(S)\end{tabular}\\ 
\hline

Baseline (IGEV) & & &  & & 0.47& 5.21 & 0.37\\ 
\hline
Mono+Conv &  \checkmark   &  Conv  & &  & 0.46 & 5.12 & 0.64  \\
Mono+MGR &   \checkmark  &  MGR  &  & & 0.43  & 4.96 & 0.65 \\
Mono+MGR+Conv &  \checkmark  & MGR &  Conv &  & 0.42 & 4.82 & 0.65 \\ 
Mono+MGR+SGA &  \checkmark  & MGR &  SGA &  & 0.39 & 4.43 & 0.66\\  
Full model (MonSter++) &  \checkmark  & MGR &  SGA & \checkmark & \textbf{0.37} & \textbf{4.25} & 0.64\\ 

\hline
\end{tabular}
\label{tab:abla}
\end{table*}

\subsection{Ablation Study}
\label{sec:ablation}
To demonstrate the effectiveness of each component of MonSter++,  we ablation on Scene Flow~\cite{dispNetC} and the following strategies are discussed:
\begin{itemize}
    \item \textbf{Disparity Fusion}: Monocular depth often carries substantial noise, which complicates its fusion with the stereo branch’s disparity. To demonstrate the efficiency of our Mono Guided Refinement (MGR) fusion method, we replaced the MGR module with a convolution-based hourglass network of equal parameter numbers, directly concatenating the monocular and stereo disparities for fusion, denoted as Mono+Conv. As shown in Tab.\ref{tab:abla}, compared with `Mono+Conv', `Mono+MGR' improves the EPE metric by 6.52\%.
    \item \textbf{Scale$\And$Shift Refinement}: As shown in Fig.\ref{fig:align_vis}, even after global scale-shift alignment, the monocular depth still suffers from significant scale and shift ambiguities. Therefore, leveraging high-confidence regions from the stereo branch to correct the scale and shift of the monocular disparity is essential. We incorporated our Stereo Guided Refinement module (SGA) to optimize the scale and shift based on `Mono+MGR'. As shown in Tab.\ref{tab:abla}, compared with `Mono+MGR', our SGA module achieves improvements of 9.30\% in EPE and 10.69\% in 1-pixel error, demonstrating that monocular depth with accurate scale and shift can provide further enhancements to stereo matching. Similarly, we validated that our SGA module is more effective than direct convolution-based fusion. Compared to `Mono+MGR+Conv', our SGA still achieves an 8.09\% improvement in 1-pixel error.
    \item \textbf{Feature Sharing}: The feature extraction component of monocular depth estimation models, pre-trained on large datasets, contains rich contextual information that can be shared with the stereo branch. We employ a feature transfer network to map these features, transforming them into a representation compatible with stereo matching. As shown in the last 2 rows of Tab.\ref{tab:abla}, feature sharing improved the EPE metric by 5.13\%.
\end{itemize}

\textbf{Compatibility with Other Monocular Models.} To validate the generalizability of MonSter++ to different depth models, we replace the monocular branch with the ViT-large version of \cite{depthany_v1} and the dpt\_beit\_large version of MiDaS \cite{midas}. As shown in Tab.\ref{tab:abla2}, all MonSter++ variants consistently outperform the baseline \cite{igev}. This demonstrates the versatility of MonSter++ and its potential to benefit from future advancements in monocular depth models.

\begin{table}[t] \footnotesize
  \centering
  \caption{
  \textbf{Efficiency and universality of MonSter++.} The \textbf{best} result is bolded.}
  \setlength{\tabcolsep}{1.75pt}
  \begin{tabular}{lcccc}
    \toprule
    Model & \begin{tabular}[c]{@{}c@{}}Mono Depth\\   Model\end{tabular} & \begin{tabular}[c]{@{}c@{}}Iteration \\ Number\end{tabular} & \begin{tabular}[c]{@{}c@{}}Sceneflow \\ (EPE)\end{tabular}& \begin{tabular}[c]{@{}c@{}}Run-time\\(S)\end{tabular}
    \\
     \midrule
    IGEV\cite{igev} & & 32  & 0.47 & 0.37\\  
    Full model & DepthAnythingV2\cite{depthany_v2} & 32  & \textbf{0.37} & 0.64 \\
    Full model-4iter & DepthAnythingV2\cite{depthany_v2} & 4  & 0.42 & \textbf{0.34} \\
    \hline
    Full model-V1 & DepthAnythingV1\cite{depthany_v1} & 32  & 0.39 & 0.64 \\
    Full model-midas & MiDaS\cite{midas} & 32  & 0.41 & 0.51 \\

    \bottomrule
  \end{tabular}
  \label{tab:abla2}
\end{table}

\textbf{Efficiency.} We use the ViT-large version of DepthAnythingV2 \cite{depthany_v2} as the monocular branch, which introduces time and memory overheads. Compared to the baseline~\cite{igev}, our inference time increases from 0.37s to 0.64s. With the monocular branch containing 335.3M parameters, the stereo branch having 12.6M parameters, and the SGA and MGR modules having 8.2M parameters, our full model has a total of 356.1M parameters. Existing methods \cite{madis, croco} also use ViT-based encoders with a large number of parameters. For example, CroCo-Stereo \cite{croco} has 437.4M parameters, larger than MonSter++, yet its performance is inferior, as shown in Tab.\ref{tab:benchmark}, this demonstrates that the number of parameters does not determine the final effectiveness. Our research focuses on the accuracy and generalization capabilities, given the significant improvement in accuracy and generalization (as shown in Tab.\ref{tab:benchmark}, Tab.\ref{tab:zero-all} and Tab.\ref{tab:generalization}), the additional inference cost is acceptable. Notably, our approach simplifies stereo matching and enables us to achieve SOTA performance with only 4 iterations ($N_{1}$=$N_{2}$=2). As shown in Tab.\ref{tab:abla2}, while the baseline requires 32 iterations, our method achieves a 10.64\% improvement in EPE with just 4 iterations, achieving a better accuracy-speed trade-off. Further memory reduction can be achieved through encoder quantization or distillation, which we leave as future work.

\section{Conclusion}
We propose MonSter++, a novel pipeline that decouples the multi-view depth estimation into a simpler paradigm of recovering scale and shift from relative depth. This approach fully leverages the contextual and geometric priors provided by monocular methods while avoiding issues such as noise and scale ambiguity.  Consequently, our approach markedly improves both the accuracy and robustness of depth estimation, particularly in ill-posed regions. MonSter++ achieves substantial gains over state-of-the-art methods across eight widely used benchmarks spanning three tasks—stereo matching, real-time stereo matching, and multi-view stereo. We also introduce a real-time version, RT-MonSter++, which delivers the best accuracy among all published real-time methods. To foster highly generalizable depth models for the open-source community, we conduct large-scale training on 2 million image pairs for both MonSter++ and RT-MonSter++. All models and training code will be publicly released.

\noindent\textbf{Acknowledgement.} Authors from HUST are supported by National Natural Science Foundation of China(62472184) and the Fundamental Research Funds for the Central Universities.

\bibliographystyle{IEEEtran}
\bibliography{main}

\begin{IEEEbiography}[{\includegraphics[width=1in,height=1in,clip,keepaspectratio]{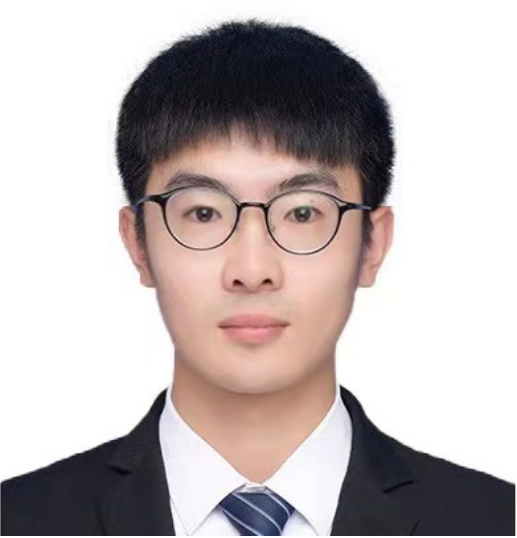}}]{Junda Cheng}
  is a PhD student at the Department of Electronic Information and Communications at Huazhong University of Science and Technology. He is supervised by Prof. Xin Yang. He received his B.Eng. degree from Huazhong University of Science and Technology in 2020. His research interests include stereo matching,  multi-view stereo, and deep visual odometry. He has published multiple papers in IEEE-TPAMI, IJCV, AAAI, and CVPR.
\end{IEEEbiography}

\begin{IEEEbiography}[{\includegraphics[width=1in,height=1in,clip,keepaspectratio]{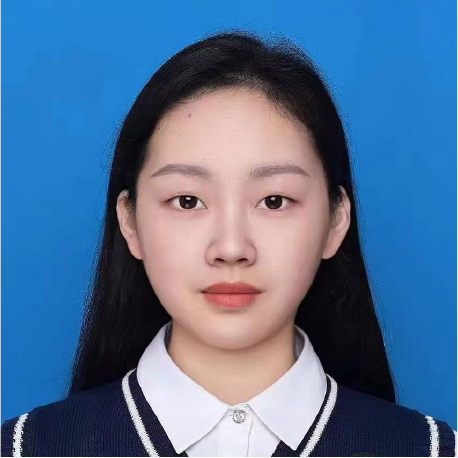}}]{Wenjing Liao}
  is a master's student at the School of Electronic Information and Communications at Huazhong University of Science and Technology. She is supervised by Prof. Xin Yang. She received her B.Eng. degree from Huazhong University of Science and Technology in 2025. Her research focuses on stereo matching and multi-view stereo.
\end{IEEEbiography}

\begin{IEEEbiography}[{\includegraphics[width=1in,height=1in,clip,keepaspectratio]{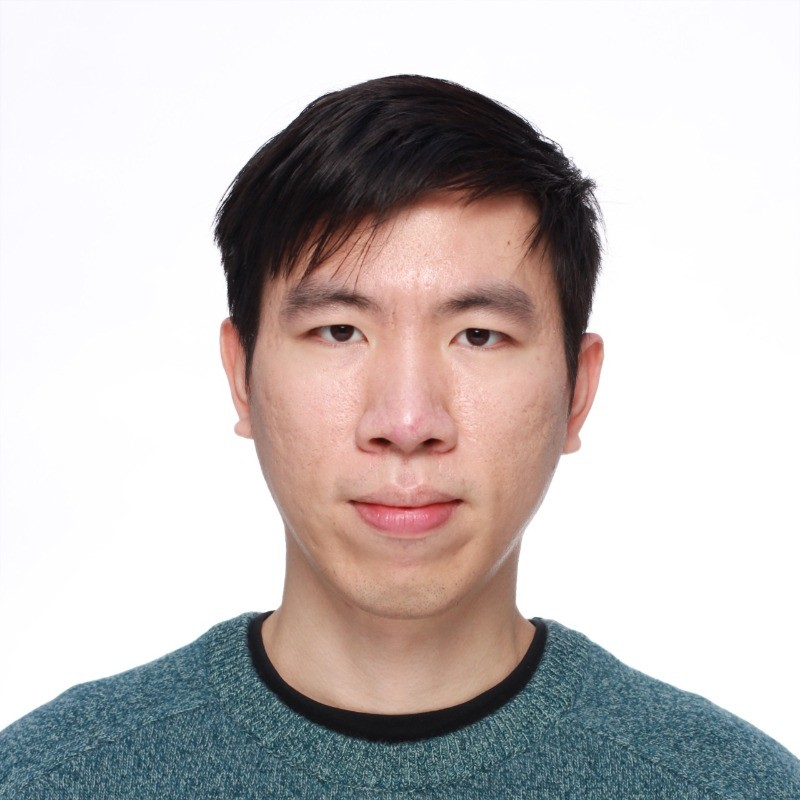}}]{Zhipeng Cai} is a Senior Researcher at Meta. He received the PhD degree in computer science from the University of Adelaide, Australia. His research interests include perception, optimization, and vision language models. Several of his representative works have been selected as top-tier conference oral/spotlights. One of them has been selected as one of the 12 best papers at ECCV18.
\end{IEEEbiography}

\begin{IEEEbiography}[{\includegraphics[width=1in,height=1in,clip,keepaspectratio]{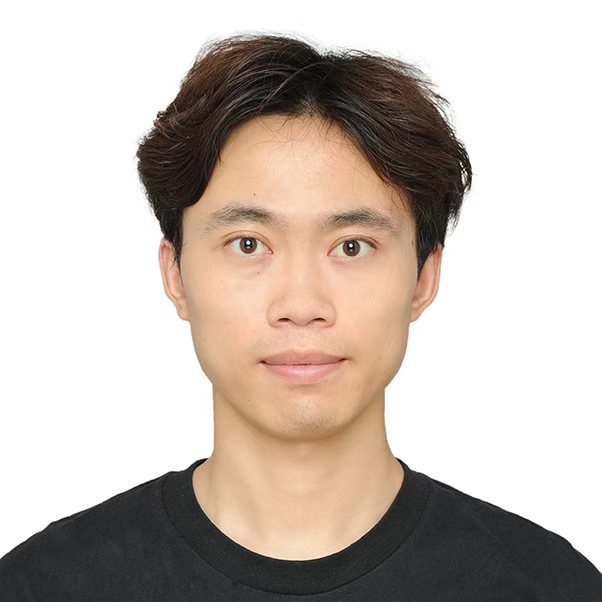}}]{Longliang Liu}
  is a master’s student in the Department of Electronic Information and Communications at Huazhong University of Science and Technology. He is supervised by Prof. Xin Yang. He obtained his B.Eng. degree from Huazhong University of Science and Technology in 2024. His areas of research interest encompass optical flow, stereo matching, and multi-view stereo. He has authored several papers in venues including CVPR, ICCV, and 3DV.
\end{IEEEbiography}

\begin{IEEEbiography}[{\includegraphics[width=1in,height=1in,clip,keepaspectratio]{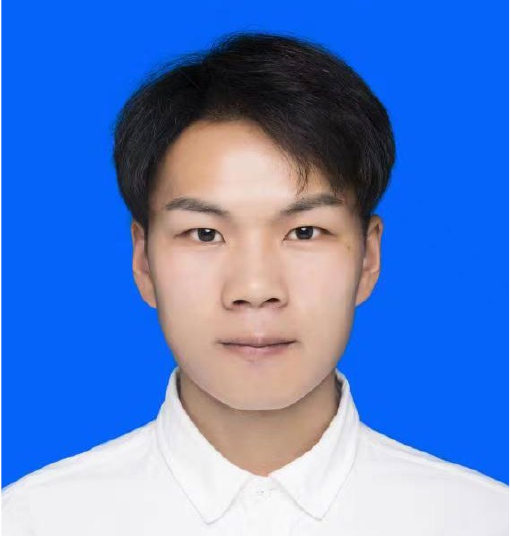}}]{Gangwei Xu}
  is a PhD student at the Department of Electronic Information and Communications at Huazhong University of Science and Technology. He is supervised by Prof. Xin Yang. He received his B.Eng. degree from Huazhong University of Science and Technology in 2021. His research interests include stereo matching, optical flow estimation and HDR video reconstruction. He has published multiple papers in IEEE-TPAMI, IJCV, NeurIPS, ICCV, and CVPR.
\end{IEEEbiography}

\begin{IEEEbiography}[{\includegraphics[width=1in,height=1in,clip,keepaspectratio]{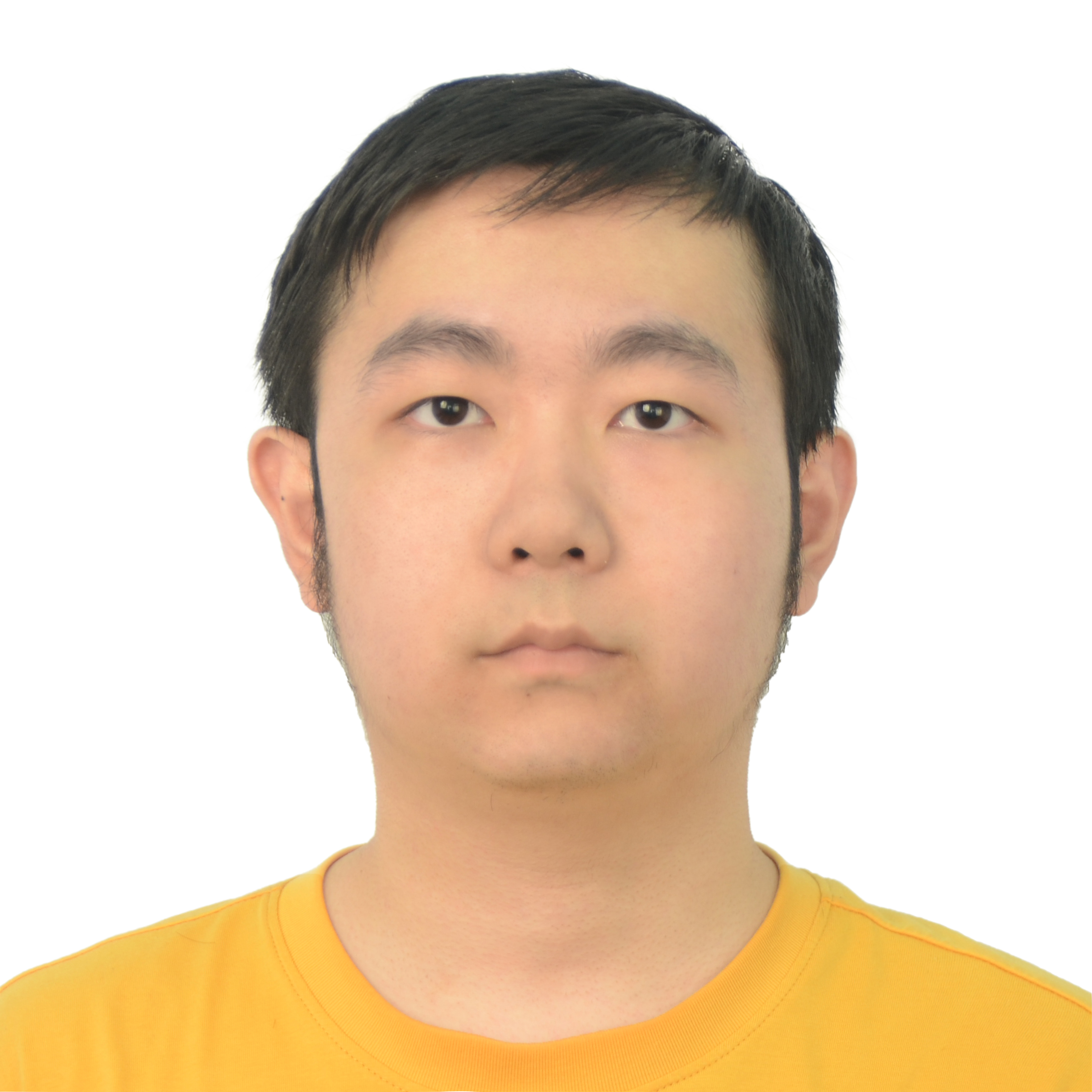}}]{Xianqi Wang}
  is a PhD student at the Department of Artificial Intelligence and Automation at Huazhong University of Science and Technology. He is supervised by Prof. Xin Yang. He received his B.Eng. degree from Huazhong University of Science and Technology in 2022. His research interests include stereo matching and multi-view stereo. He has published multiple papers in IEEE-TPAMI, ICCV, and CVPR.
\end{IEEEbiography}

\begin{IEEEbiography}[{\includegraphics[width=1in,height=1in,clip,keepaspectratio]{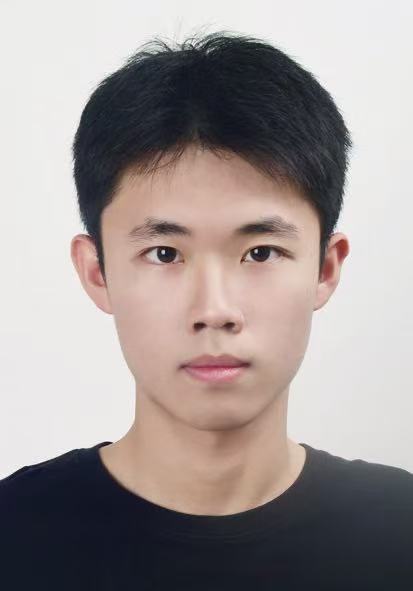}}]{Yuzhou Wang}
  is a master's student at the School of Electronic Information and Communications at Huazhong University of Science and Technology. He is supervised by Prof. Xin Yang. He received her B.Eng. degree from Huazhong University of Science and Technology in 2025. His research focuses on stereo matching and multi-view stereo.
\end{IEEEbiography}

\begin{IEEEbiography}[{\includegraphics[width=1in,height=1in,clip,keepaspectratio]{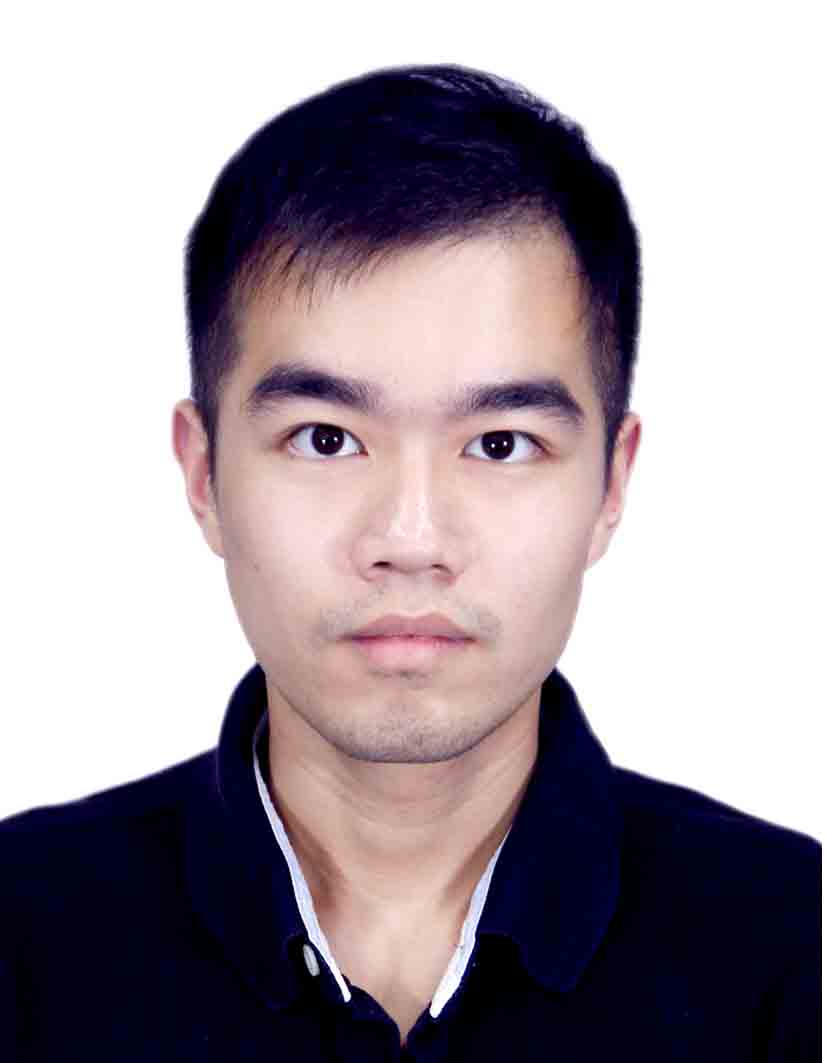}}]{Zikang Yuan}
 is currently a postdoctoral fellow in AI Chip Center, Hong Kong University of Science and Technology (HKUST). He received his PhD degree from Huazhong University of Science and Technology (HUST), Wuhan, China, in 2024. He has published two papers on RA-L, one paper on TPAMI, three papers on IROS and one paper on ICRA. His research interests include SLAM, neural rendering and autonomous exploration.
\end{IEEEbiography}

\begin{IEEEbiography}[{\includegraphics[width=1in,height=1in,clip,keepaspectratio]{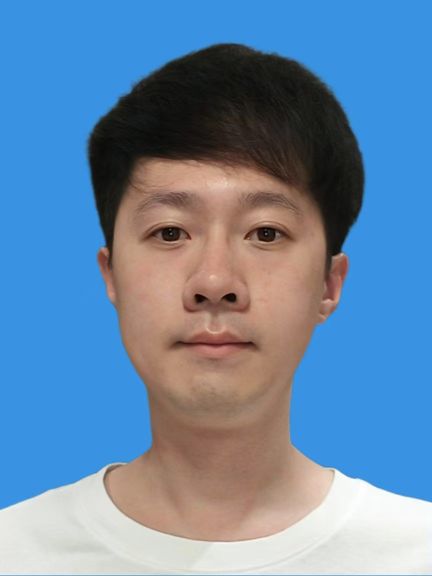}}]{Yong Deng} is a Senior Engineer at Autel Robotics. He received the PhD degree in Electrical and Computer Engineering from the National University of Singapore. His research interests include stereo matching, multi-view stereo, and semantic segmentation. He has published papers in IEEE-TIP and IEEE-TMM.
\end{IEEEbiography}

\begin{IEEEbiography}[{\includegraphics[width=1in,height=1in,clip,keepaspectratio]{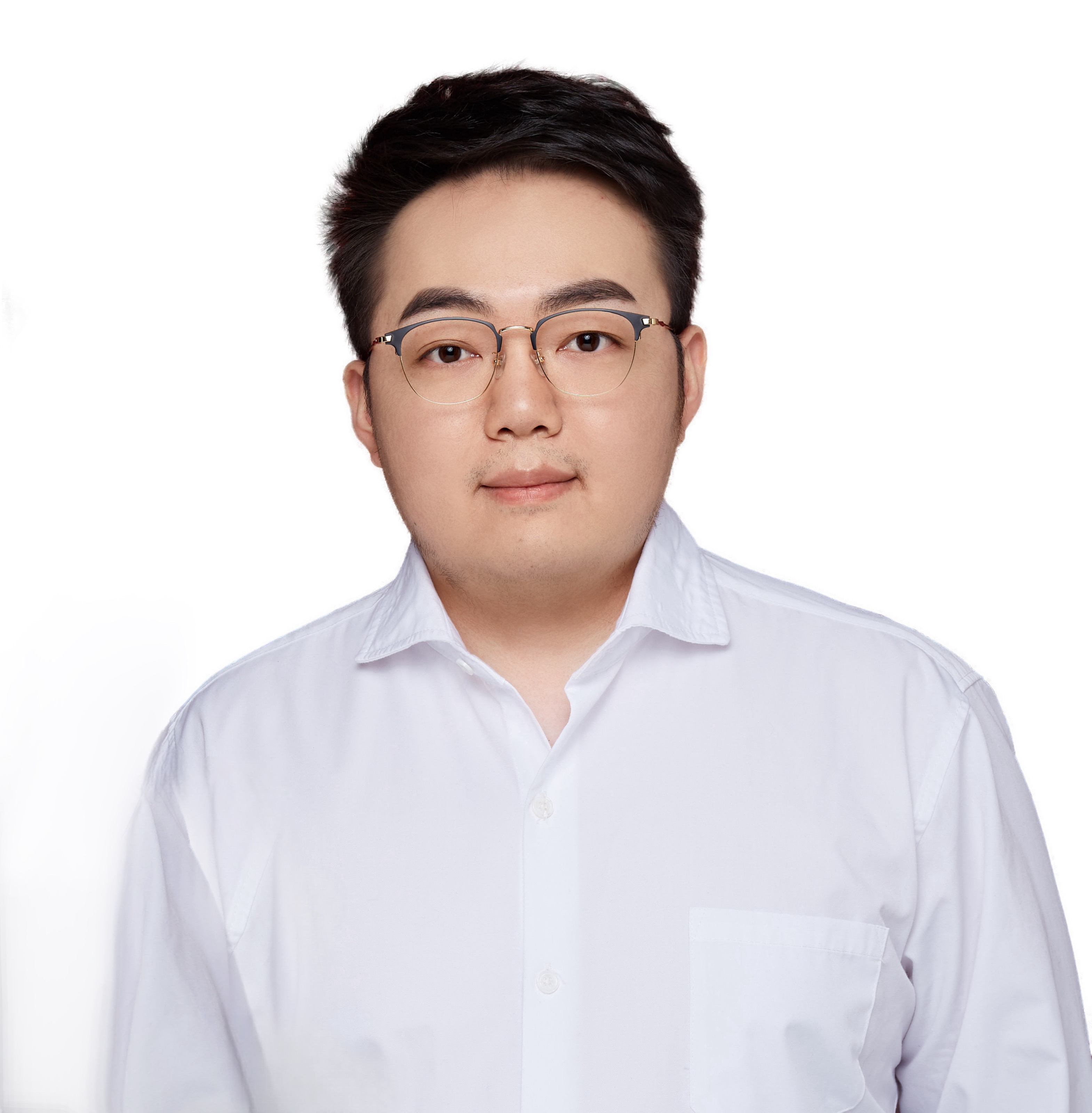}}]{Jinliang Zang} is a Senior Engineer at Autel Robotics. He received the master's degree from Xi'an Jiaotong University in Xi'an, China. His interests include stereo matching, image segmentation, and pose estimation. His works have been implemented in the drone field.
\end{IEEEbiography}

\begin{IEEEbiography}[{\includegraphics[width=1in,height=1in,clip,keepaspectratio]{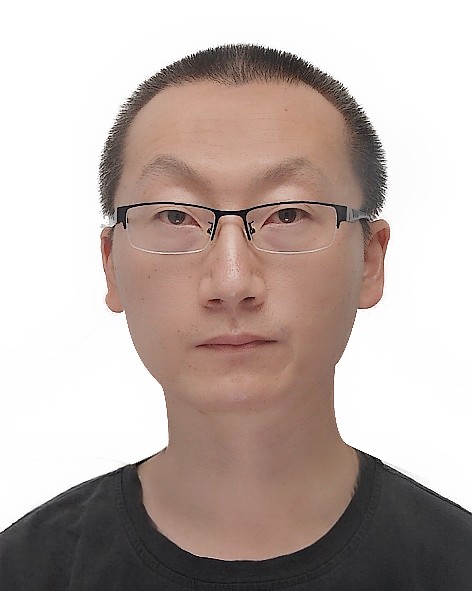}}]{Yangyang Shi}
   is a Senior Staff Research Scientist Manager at Meta and an IEEE Senior Member. He received his PhD from the Faculty of Electrical Engineering, Mathematics, and Computer Science at Delft University of Technology. His research focuses on large language models (LLMs) and generative AI, with an emphasis on applying these technologies to enhance a range of devices, including mobile phones, smartwatches, smart speakers, and AR/VR systems. Additionally, he serves as a member of the IEEE SPS Speech and Language Processing Technical Committee and a reviewer for several top tier conferences.
\end{IEEEbiography}

\begin{IEEEbiography}[{\includegraphics[width=1in,height=1in,clip,keepaspectratio]{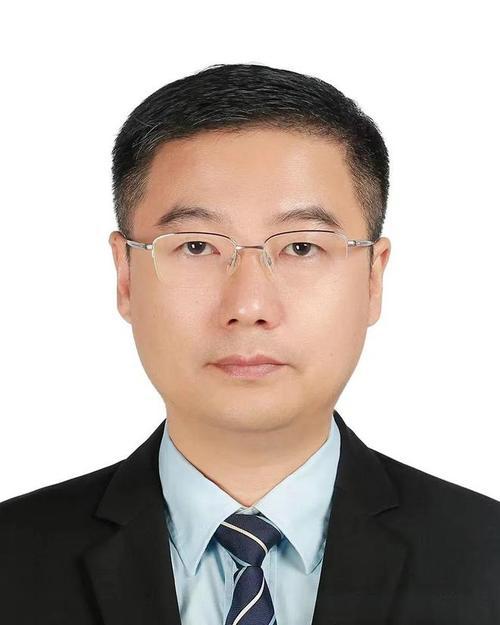}}]{Jinhui Tang}
  is a Professor at the Nanjing Forestry University. He received the B.Eng. and Ph.D. degrees from the University of Science and Technology of China in 2003 and 2008, respectively. He has authored over 150 papers in top-tier journals and conferences, with more than 10,000 citations in Google Scholar. His research interests include multimedia analysis and computer vision. He was a recipient of the best paper awards in ACM MM 2007, PCM 2011 and ICIMCS 2011, the Best Paper Runner-up in ACM MM 2015, and the best student paper awards in MMM 2016 and ICIMCS 2017. He has served as an Associate Editor for the IEEE TNNLS, IEEE TKDE, IEEE TMM and IEEE TCSVT.
\end{IEEEbiography}

\begin{IEEEbiography}[{\includegraphics[width=1in,height=1in,clip,keepaspectratio]{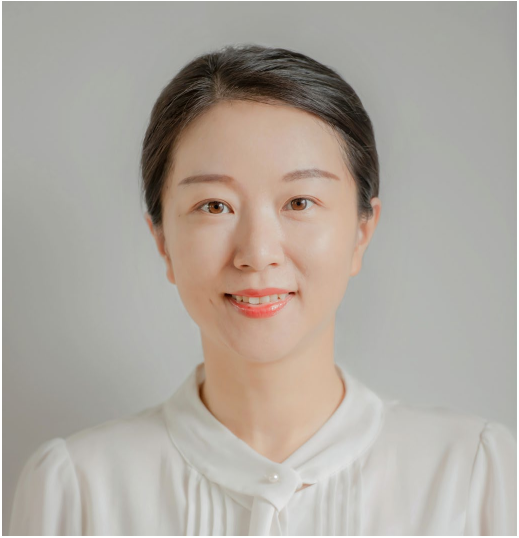}}]{Xin Yang}
   is a Professor at the Department of Electronic Information and Communications at Huazhong University of Science and Technology. She received her Ph.D. degree in the Department of Electrical Computer Engineering at the University of California, Santa Barbara (UCSB).  Her research interests include medical image analysis and 3D vision. She is the recipient of the National Natural Science Fund of China for Excellent Youth Scholar and China Society of Image and Graphics Qingyun Shi Female Scientist Award. She has published over 90 technical papers and holds 20 patents. She serves as an Associate Editor of IEEE-TVCG, IEEE-TMI, and Multimedia System, an Area Chair of CVPR’24, MICCAI’19-21, and ACM MM’18. She is also a reviewer of top journals such as IEEE-TPAMI, IJCV, etc.
\end{IEEEbiography}

\vfill

\end{document}